%% file: Author_tex.tex
\documentclass{comnet}

\usepackage{todonotes}
\usepackage{xspace}
\usepackage{caption}
\usepackage{subcaption}
\usepackage{tikz}
\usepackage{tkz-graph}
\usepackage{graphicx}
\usepackage{amsmath}
\usepackage{amssymb}
\usepackage{url}
\usepackage{booktabs}
\usepackage{lipsum}
\usepackage{soul}
\usepackage{colortbl}

\definecolor{Burgundy}{RGB}{144, 0, 32}
\definecolor{ForestGreen}{HTML}{014421}

\newcolumntype{u}{>{\columncolor{Burgundy!20}}c}
\newcolumntype{y}{>{\columncolor{cherryblossompink!20}}c}
\newcolumntype{n}{>{\columncolor{celadon!20}}c}
\newcolumntype{t}{>{\columncolor{teal!20}}c}

\input{commands}

\sethlcolor{white}

\begin{document}

\title{From Communities to Interpretable Network and Word Embedding: an Unified Approach}

\shorttitle{From Communities to Interpretable Network and Word Embedding} 
\shortauthorlist{T. Prouteau, N. Dugué, S. Guillot} 

\author{
\name{Thibault Prouteau and Nicolas Dugué$^*$}
\address{Université du Mans, Laboratoire d'Informatique de l'Université du Mans (LIUM), avenue Olivier Messiaen, 72085 Le Mans CEDEX 9, France
\\ \email{$^*$Corresponding author: nicolas.dugue@univ-lemans.fr}}
\and
\name{Simon Guillot}
\address{Université du Mans, Laboratoire d'Informatique de l'Université du Mans (LIUM), avenue Olivier Messiaen, 72085 Le Mans CEDEX 9, France
\\ INaLCO, ERTIM, 65 rue des Grands Moulins,
75214 Paris, France}
}
\maketitle

\begin{abstract}
{Modeling information from complex systems such as humans social interaction or words co-occurrences in our languages can help to understand how these systems are organized and function. Such systems can be modeled by networks, and network theory provides a useful set of methods to analyze them. Among these methods, \textit{graph embedding} is a powerful tool to summarize the interactions and topology of a network in a vectorized feature space. When used in input of machine learning algorithms, embedding vectors help with common graph problems such as link prediction, graph matching, etc. In Natural Language Processing (\nlp), such a vectorization process is also employed. \textit{Word embedding} has the goal of representing the sense of words, extracting it from large text corpora. Despite differences in the structure of information in input of embedding algorithms, many graph embedding approaches are adapted and inspired from methods in \nlp. Limits of these methods are observed in both domains. Most of these methods require long and resource greedy training. Another downside to most methods is that they are \textit{black-box}, from which understanding how the information is structured is rather complex. Interpretability of a model allows understanding how the vector space is structured without the need for external information, and thus can be audited more easily. With both these limitations in mind, we propose a novel framework to efficiently embed network vertices in an interpretable vector space. Our \textit{Lower Dimension Bipartite Framework} (\LDBGF) leverages the bipartite projection of a network using cliques to reduce dimensionality. Along with \LDBGF, we introduce two implementations of this framework that rely on communities instead of cliques: \SINrNR and \SINrMF. 
We show that \SINrMF can perform well on classical graphs and \SINrNR can produce high-quality graph and word embeddings that are interpretable and stable across runs.}
{graph embedding; word embedding; interpretability}
\end{abstract}

\section{Introduction}

Network science provides versatile tools to model the organization of real-world systems. Instances of real-world systems that can be modeled with graphs are varied. Such systems include biological networks modeling how protein interact in a living organism, social interactions on online social networks or in workplaces, geographical organization with road and railroad networks, the organization of an online encyclopedia with connection between its pages, scientific publication systems with collaboration networks, etc. Network science and graphs provide a general framework to represent interactions between items regardless of the type of structure described, but also to extract significant information about the organization of the systems modeled. For instance, from the biological network structure, one might want to uncover groups of similar highly intertwined proteins. Furthermore, from a social network, one may predict how many friends a user might have.
Such tasks increasingly rely on machine learning algorithms that use a vectorized representation in input. This is where representation learning comes into play, what is commonly called \textit{graph embedding} or \textit{node embedding}, namely projecting nodes in a vector space that encompasses as well as possible graph topology from local, to more distant interactions and organization. Multiple methods have proposed solutions to the task of \textit{graph embedding} to automatically extract vector representations of nodes. Among the methods later detailed in Section~\ref{sec:related_work}, we can cite \FDWLK~\cite{perozzi_deepwalk_2014} and \FWKLT~\cite{perozzi_dont_2017} for random walks based approaches, \HOPE~\cite{ou_asymmetric_2016} and \FVERSE~\cite{tsitsulin_verse_2018} for matrix factorization approaches, and \FLouvainNE~\cite{bhowmick_louvainne_2020}, based on community detection. 

Although \textit{graph embedding} allows summarizing the topology of a graph in a vector, \textit{embedding} data was first popularized for words in the form of \textit{word embedding}. Word embedding aims at modeling semantic proximity from unstructured text data with a few dimensions. These representations also allow limiting the number of dimensions needed to convey the semantics of a large lexicon. Word embedding and network embedding are said to stem from the distributional hypothesis, first introduced by linguist \emph{Z. S. Harris} in 1954 with the following remark: "\textit{linguistic items with a similar distribution have a similar meaning}" \cite{harris_distributional_1954}. A few years later, in 1957, \emph{J. R. Firth} contributed to the distributional hypothesis with the famous "\textit{You shall know a word by the company it keeps}". Following this hypothesis, word embedding models rely on the context in which words appear, and graph embedding methods may rely on random walks to generate sequences of vertices that can be considered sentences. This kinship is further highlighted by the \emph{Skip-Gram} algorithm employed in \FWtoV~\cite{mikolov_efficient_2013} for word embedding, and also in \FDWLK~\cite{perozzi_deepwalk_2014} and \FWKLT~\cite{perozzi_dont_2017} for graph embedding. {Yet, graph embedding methods could also be applied to word co-occurrence networks, leveraging graph topology for representation. In such networks, vertices represent words and edges their co-occurrence. \textit{Tang et al.}}~\cite{tang_line_2015} {and \textit{Yang et al.}}~\cite{yang2015network} {introduced methods that can embed jointly from network structure and word information. Unlike} \cite{tang_line_2015,yang2015network}{, our method does not simultaneously learn graph embeddings and word embeddings but provides a unified pipeline that can derive graph embeddings and word embeddings.}

Furthermore, embedding techniques have limits, they are usually computationally expensive, parameter dependent, and lead to dense latent spaces in which dimensions are not interpretable.
{Dependency on parameters means that multiple models may need to be trained untils the optimal set of hyperparameters is found.}
{Efficiency of computation is critical to reduce the environmental impact of artificial intelligence (AI) in the context of global warming.}
It is a growing concern in academia, with studies by \emph{Strubell et al.}~\cite{strubell_energy_2020}, \emph{Lannelongue et al.}~\cite{lannelongue_green_2021} and \emph{Patterson et al.}~\cite{patterson_carbon_nodate} advocating for more consideration of the environmental impact of models. Lastly, interpretability is another critical feature to gain insight into the internal organization of the complex system described. For example, how the sense of a word is composed or what are the dynamics of collaboration in a co-authorship network. Interpretability is also a key challenge to building trustworthy AI systems~\cite{rudin_stop_2019}, especially in sensitive applications such as legal or medical \nlp~\cite{neveol2022french,kim2023race}. Interpretability is for instance, useful to uncover, characterize, and potentially mitigate biases (gender, race, age biases for instance) propagated from data, and thus enforce fairness in AI systems, such as discussed in the survey of Choudhary et al.~\cite{choudhary2022survey} for machine learning on graphs.

Based on this kinship between \textit{graph embedding} and \textit{word embedding} and the need for more efficient, and interpretable representation learning algorithms,
we introduce a \textbf{network embedding} framework that may be applied to many types of networks. In particular, we address graph and word embedding with the same, unified graph framework. We are considering word co-occurrence networks extracted from text corpora in the case of word embedding. The \emph{Lower Dimension Bipartite Framework} (\LDBGF) we introduce is a theoretical framework that relies on a bipartite projection of the network to extract the vertex embeddings. This bipartite projection is a tangible graph object, which makes it interpretable, and thus easy to audit. We then describe two implementations that fit within the \LDBGF framework, and that are based on communities in networks: \SINrNR and \SINrMF. The first one, \SINrNR is based on an \textit{ad hoc} measure of connectivity between vertices and communities. On the other hand, \SINrMF is fully unsupervised and relies on gradient descent to derive a vector space. These methods allow extracting interpretable representation directly correlated with the communities detected in the network. 

Our experiments are designed to demonstrate the capacities of models to embed information at three scales: microscopic (vertex and neighbors), mesoscopic (community) and macroscopic (whole network), and on a specific application to word co-occurrence networks.

Our contributions are the following:
\begin{itemize}
    \item A theoretical framework (\LDBGF) to embed network vertices in a sparse interpretable space. Methods within this framework may be implemented in various ways.
    
    \item Two implementations of the \LDBGF leading to interpretable vectors based on communities: \SINrNR (Node Recall) and \SINrMF (Matrix Factorization).

    \item A thorough evaluation of the performances of our implementations on classical tasks against state-of-the-art methods across network theory, showing their relevance.

    \item \SINrNR runs in linear time and does not require many computational resources: it is the most efficient approach of the two we implemented.
   
    \item An evaluation on the interpretability of \SINrNR regarding words.
    
    \item Experiments demonstrating the performance and stability of \SINrNR across trainings in comparison with other word embedding approaches.
    
\end{itemize}

\paragraph{Outline.} 
Since graph embedding methods have found inspiration in their word embedding counterparts, we introduce in Section~\ref{sec:related_work} related work on graph and word embeddings. In this paper, we focus on interpretability by design, subsequently we introduce motivations for interpretability and related methods in Section \ref{subsec:intrinsic_interpretability}. To introduce our novel graph embedding framework, in Section~\ref{sec:visual-introduction} we present the philosophy behind our framework through a series of visualizations on an airport network of the United States of America. In section \ref{sec:algorithmic_framework} we present the theoretical \LDBGF framework and two implementations of the latter based on community detection : \SINrNR and \SINrMF, a matrix factorization approach. Experiments are presented in Section \ref{sec:experiments} and divided in two parts. {In the first series of experiments, we assess the performance of our community-based approach on tasks at the microsopic, mesoscopic, and macroscopic-levels. The second series of experiments assesses the performance of our method on the specific application to word co-occurrence networks. Our final experiments in Section}~\ref{subsec:exp_robustness}{ assess the interpretability of word embeddings.} Finally, Section \ref{sec:conclusion} comments on the field of possibilities opened by \LDBGF and its implementations.

\section{Related Work}\label{sec:related_work}

\subsection{Embedding Methods}\label{subsec:embedding_methods}

\paragraph{Word embeddings.} Graph embedding methods have been heavily influenced by algorithms developed in \nlp~\cite{perozzi_deepwalk_2014}. Word embedding aims at uncovering a latent vector space in which to project words, providing a more synthetic representation of a word than a co-occurrence matrix (with a lower dimension)  while  encompassing semantic information. These representations are then used as input by classifiers, mostly neural architectures, to solve various tasks such as \textit{named entity recognition}, \textit{part-of-speech-tagging}, \textit{sentiment analysis} or \textit{machine translation}.

The first approaches to train word embeddings were the matrix factorization based. Those are directly connected to the distributional hypothesis: they factorize the word co-occurrence matrix. The literature usually defines co-occurrences inside a corpus using a sliding window parameterized by a certain size. All words within the window are said to co-occur. In the case of \nnse~\cite{murphy_learning_2012}, the co-occurrence matrix is factorized using sparse coding to enforce interpretability. For \Fglo~\cite{pennington_glove_2014}, the log co-occurrence is factorized, and the loss function is weighted by the co-occurrence frequency. Levy et al. applied \textit{Singular Value Decomposition} (\svd) to word co-occurrence matrices after having applied Positive Mutual Information on it to consider the significance of these co-occurrences~\cite{LGD15}. These methods perform well on similarity and analogy benchmarks and have been popular representations as input to a wide range of machine learning systems such as classifiers or translation systems. 

With \FWtoV, \emph{Mikolov et al.}~\cite{mikolov_efficient_2013} consider a different approach: the task of training word embedding is seen as self-supervised. Indeed, they train a logistic regression on the dot product of word embeddings: the sigmoid is supposed to be close to 1 when words co-occur, and to 0 when they do not.
This method has drawn a lot of attention and was extended to improve its robustness. For instance, \texttt{fastText} is based on sub-words and allows extracting word embeddings for words that have not been encountered in the training corpus~\cite{bojanowski_enriching_2017}. These methods lack flexibility to represent polysemous words, since they provide a single vector per type, \textit{Tian et al.}~\cite{tian2014probabilistic} introduced a multi-prototype approach providing one vector per sense. Such multi-prototype approaches paved the way for contextualized representations that provide one vector per occurrence.

Language models such as \emph{Devlin et al.}'s \bert~\cite{devlin_bert_2018} with its self-supervised transformer architecture have gained in popularity since they provide such contextualized representations. Transformers architectures implement self-attention mechanisms, where words with similar representations in the same sequence are considered to attend each other. The representation of a word is then a mixture of its embedding and of the embeddings of the words attending to it, allowing its contextualization. Training such architectures is usually performed by predicting masked words from their contexts, or predicting the next sentence. 
Subsequent transformer-based models adapted from \bert include \roberta~\cite{liu_roberta_2019} which alters key hyperparameters and training objectives, such as removing next-sentence prediction; both \tfive~\cite{raffel_exploring_2020} and \gpt~\cite{brown_language_2020, neelakantan_text_nodate} use large transformer architectures to build what is now named \textit{Large Language Models} (LLM).

\paragraph{Node embedding.} Node embedding approaches try to embed the local neighborhood of a node and nodes with similar roles in the graph closely together. They are frequently evaluated on classical tasks such as: \textit{link prediction}, \textit{node classification} or \textit{graph reconstruction}. Earliest approaches to network embeddings applied dimension reduction techniques such as \isomap with multidimensional scaling (MDS) \cite{tenenbaum_global_2000}, \lle~\cite{roweis_nonlinear_2000} or \lapeigenmap~\cite{belkin_laplacian_2001}. Then, graph embedding fed from advances made to methods in \nlp. \FWtoV has influenced the field, leading to methods that adopt random-walk sampling strategies with the \emph{Skip-Gram} model, namely \FDWLK~\cite{perozzi_deepwalk_2014}, \FWKLT~\cite{perozzi_dont_2017} and \FNtoV~\cite{grover_node2vec_2016}. It is worth noting that the \emph{Skip-Gram} model is implicitly related to matrix factorization of a word-context matrix \cite{mikolov_efficient_2013, levy_neural_2014}. Multiple approaches make use of matrix factorization to obtain node representations: \lineemb~\cite{tang_line_2015} optimizes for first and second order proximity and can jointly extract embeddings from network structure and text; \grarep~\cite{cao_grarep_2015} factorizes k-step transition matrices between nodes; \emph{Liu et al.}~\cite{liu_online_2016} extend \emph{non-negative matrix factorization} (\FNMF) to run online and scale to larger graphs; \HOPE~\cite{ou_asymmetric_2016} factorizes multiple matrices of similarity measures to preserve high-order proximity in the network; \GVNR \cite{brochier_global_2019} is an adaptation of \Fglo to node embedding that better handles non-co-occurrence of items; finally, \FVERSE~\cite{tsitsulin_verse_2018} tries to reconstruct the distribution of a chosen similarity measure for each vertex using a single layer neural network.

Some algorithms focus specifically on preserving community structures: \emph{Wang et al.}~\cite{wang_community_2017} adapt \FNMF to incorporate community structures' representations; \emph{Rozemberczki et al.}~\cite{rozemberczki_gemsec_2019} introduce \FGEMSEC combining \emph{Skip-Gram} objectives with vertex clustering to preserve community information. Not with the specific goal of preserving communities, but still using community structures.
\emph{Bhomwick et al.}~\cite{bhowmick_louvainne_2020} leverage the hierarchical structure of the \louvain community detection algorithm to learn embeddings of subgraphs in each level and afterward combine these representations to derive a vertex representation. 

Alike the emergence of neural models in \nlp, neural approaches related to \emph{Graph-Convolutional Networks} (GCN) and \emph{Graph Neural Networks} (GNN) appeared to deal with graph embedding. Notably, \SDNE~\cite{wang_structural_2016} makes use of autoencoders to preserve first- and second-order proximity; \DNGR~\cite{cao_deep_2016} also uses deep autoencoders to capture non-linearity in graph and embed nodes. Neural methods are described more extensively in the survey by \emph{Makarov et al.}~\cite{makarov_survey_2021}.

\paragraph{Computationally expensive and not interpretable.} Despite high enthusiasm of the community, neural models need larger amounts of data to train ever-growing models. Although not needing to be retrained from scratch, large transformers models (340M parameters for \bert, 11B for \tfive and 175B for \gpt) still need to be trained and tuned once on a large corpus before they can be fine-tuned on task or domain-specific data. Their training has a significant environmental impact. \emph{Strubell et al.}~\cite{strubell_energy_2020} 
draw attention to the key impacts of the race towards ever-larger pretrained models: training large language models emits large amounts of CO$_2$ from the energy consumed and requires access to compute unavailable to some researchers. \emph{Lannelongue et al.}~\cite{lannelongue_green_2021} and \emph{Patterson et al.}~\cite{patterson_carbon_nodate} introduce methods to compute the carbon footprint of \nlp models and formulate recommendations to gain efficiency both in terms of implementation and computing infrastructure. In this line of work, we focus on designing low compute approaches to solve both word and graph embedding tasks.

Furthermore, neural and matrix factorization approaches produce representation spaces that are not interpretable. To counter this lack of interpretability, multiple methods were developed to investigate models' decisions. \FLIME~\cite{ribeiro_why_2016} is a surrogate model to explain the results of a classifier, \FSHAP~\cite{lundberg_unified_2017} proposes a method to analyze the contribution of individual features to a decision. These methodologies can outline what led a model to a prediction, but in a post-hoc manner—on top of the model audited or once it has been extracted. Post-hoc interpretability is a step in the right direction. However, the model explained is still intrinsically a black box. Furthermore, it is another model to train explaining the first one, making this solution more expensive to compute. Models with interpretability by design rather than post-hoc allow interpretability of the representation without the need for an additional model to generate explanations of its internal logic. This is what differentiates interpretable methods from explainable methods. According to \emph{Rudin et al.}~\cite{rudin_stop_2019}, interpretable methods should be preferred where possible. Following this distinction between interpretable and explainable methods, we define interpretable approaches as methods whose dimensions can be audited by design without post-hoc processing. We detail in the next subsection the related work considering interpretability.

\subsection{Model-Intrinsic Interpretability}\label{subsec:intrinsic_interpretability}

As stated by \emph{Rudin et al.}~\cite{rudin_stop_2019}, interpretability of a system is often defined by opposing it to explainability, which is considered as the post-hoc explanation of a system's decisions. But some authors also define interpretability for systems as their ability to produce meaningful outputs understandable by non-expert users~\cite{broniatowski_psychological_2021}. Both of these definitions are complementary, and we embrace both of them in this work.

Interpretability of an embedding space is commonly defined in the literature \cite{murphy_learning_2012, subramanian_spine_2018} as the capacity for humans to make sense of the dimensions in the embedding space produced. These dimensions can be seen as themes, described by a consistent set of words. These dimensions can thus be seen as semantic features of words, each word being represented by a small set of these features. Most embedding methods do not provide interpretable embedding spaces according to this definition. There is a common denominator to most interpretable methods, as first introduced with \nnse~\cite{murphy_learning_2012}:
\begin{enumerate}
    \item high dimensionality of the latent space uncovered. It is motivated by the difficulty to represent the meaning of a large lexicon about many distinct topics with only a few dimensions, and thus a few semantic features.
    \item sparsity of the vectors and dimensions. This is directly connected to the high-dimensionality property. To have dimensions that can act as semantic features, these dimensions should be semantically consistent, and thus only a subset of the vocabulary is supposed to take part in this dimension, leading to sparseness.
    \item non-negativity of the values. It is not computationally efficient to store negative features alongside positive ones, and psycholinguistic experiments show that it is also not cognitively efficient~\cite{PALMER1977441,lee99}.
\end{enumerate}

Early on, \emph{Murphy et al.}~\cite{murphy_learning_2012} introduce an interpretable word embedding model, \emph{Non-negative sparse embedding} (\nnse) which factorizes a word co-occurrence matrix, producing a 1000-dimension vector space. This space is bigger than the classic $300$-dimension models for \FWtoV. Furthermore, as the consequence of a $l1$ regularization, \nnse embeddings are sparse. \emph{Faruqui et al.}~\cite{faruqui_sparse_2015} later introduced \textit{Sparse Overcomplete Word Vectors} (\FSPOWV) that builds on the improvements achieved by \Fglo and \FWtoV. In concrete terms, from pretrained \FWtoV or \Fglo vectors, \FSPOWV applies regularizations to sparse code the vectors in an overcomplete space. Dimension of the resulting space is larger than that of the pretrained vectors, from 300 dimensions for the pretrained vectors to 500-3,000 dimensions in the resulting space. \emph{Subramanian et al.}'s \FSPINE~\cite{subramanian_spine_2018} also derives sparse interpretable word embeddings from pretrained vectors using sparse auto-encoders with losses specifically enforcing sparsity. \emph{Panigrahi et al.}~\cite{panigrahi_word2sense_2019} introduce \FWtoS, a generative model based on \emph{Latent Dirichlet Allocation} extracting dimensions that act as senses and represents  words as a probability distribution over these senses. 

Regarding node embeddings, interpretability is also a concern: \emph{Duong et al.}~\cite{duong_interpretable_2019} use a min-cut loss to uncover thematic groups of nodes in graphs and \FiGNN introduced in \emph{Serra et al.}~\cite{serra_interpreting_2021}. A method conjointly to learn a node embedding along with a textual explanation. Authors of this paper introduced a new framework for interpretable node embedding: \textit{Lower Dimension Bipartite Graph Framework} (\LDBGF)~\cite{prouteau_sinr_2021}. It is a theoretical framework relying on bipartite representations of graphs to uncover a latent representation space. The first implementation of this framework in \emph{Prouteau et al.}~\cite{prouteau_sinr_2021} leverages communities, efficiently detected using \emph{Blondel et al.}'s \louvain algorithm~\cite{blondel_fast_2008}, and relations between nodes and communities to derive sparse interpretable node representations. Each dimension of the vector space is related to a dense community of nodes that should exhibit similarities. \SINrNR and \SINrMF that we introduce in this paper also implement \LDBGF, and are thus in keeping with other methods leveraging communities \cite{wang_community_2017, rozemberczki_gemsec_2019, bhowmick_louvainne_2020, duong_interpretable_2019}, while focusing on using communities for interpretability \cite{prouteau:hal-03770444}. For textual data, it is in particular in direct filiation with the method introduced by Chen et al.~\cite{chen_unsupervised_2008} that leverages community detection in keyword co-occurrence networks to extract meaning.

Our model is part of the effort to propose models requiring less computational power and more interpretability. Despite less visible than neural models, which have become rather ubiquitous and become larger and larger, there is still a whole body of research aiming for alternative interpretable latent spaces \cite{rozemberczki_gemsec_2019, duong_interpretable_2019, prouteau_sinr_2021, murphy_learning_2012, faruqui_sparse_2015, subramanian_spine_2018}. Since our focus is on interpretability of the model while keeping cost and complexity of computation low, we will not be considering deep neural network approaches relying on transformers and graph-convolutional neural networks in the remainder of this work.

The main advantage of interpretability is the opportunity to visualize how information flows and is structured in a model. As a way to get across the intuition behind our \LDBGF framework and its \SINrNR and \SINrMF implementations, we start with a simple use case: representing airports and domestic connections within the United States.

\section{A visual introduction: embedding U.S. airports}
\label{sec:visual-introduction}

\paragraph{}To present the philosophy behind \LDBGF that we will more formally introduce Section~\ref{sec:algorithmic_framework}, we first consider a use case based on an airport network of domestic flights in the United States of America. From a graph standpoint, we can represent the network of airports in multiple ways. The simplest way to build such a graph is to connect any two airports between which a direct route exists. The graph can be weighted in numerous manners according to the number of daily flights, the number of passengers or the distance between airports. However, for the sake of simplicity and because most plane rotations are bidirectional, we consider an undirected and unweighted graph $G=(V,E)$ of US airports with $V$ the set of vertices representing the airports, $E$ the set of edges representing the existence of a flight between two airports. Graph $G$ is drawn over a map of the USA in Figure \ref{fig:routes} and shows the sheer number of domestic connections between airports.
\begin{figure}[h]
    \centering
    \includegraphics[width=.7\textwidth]{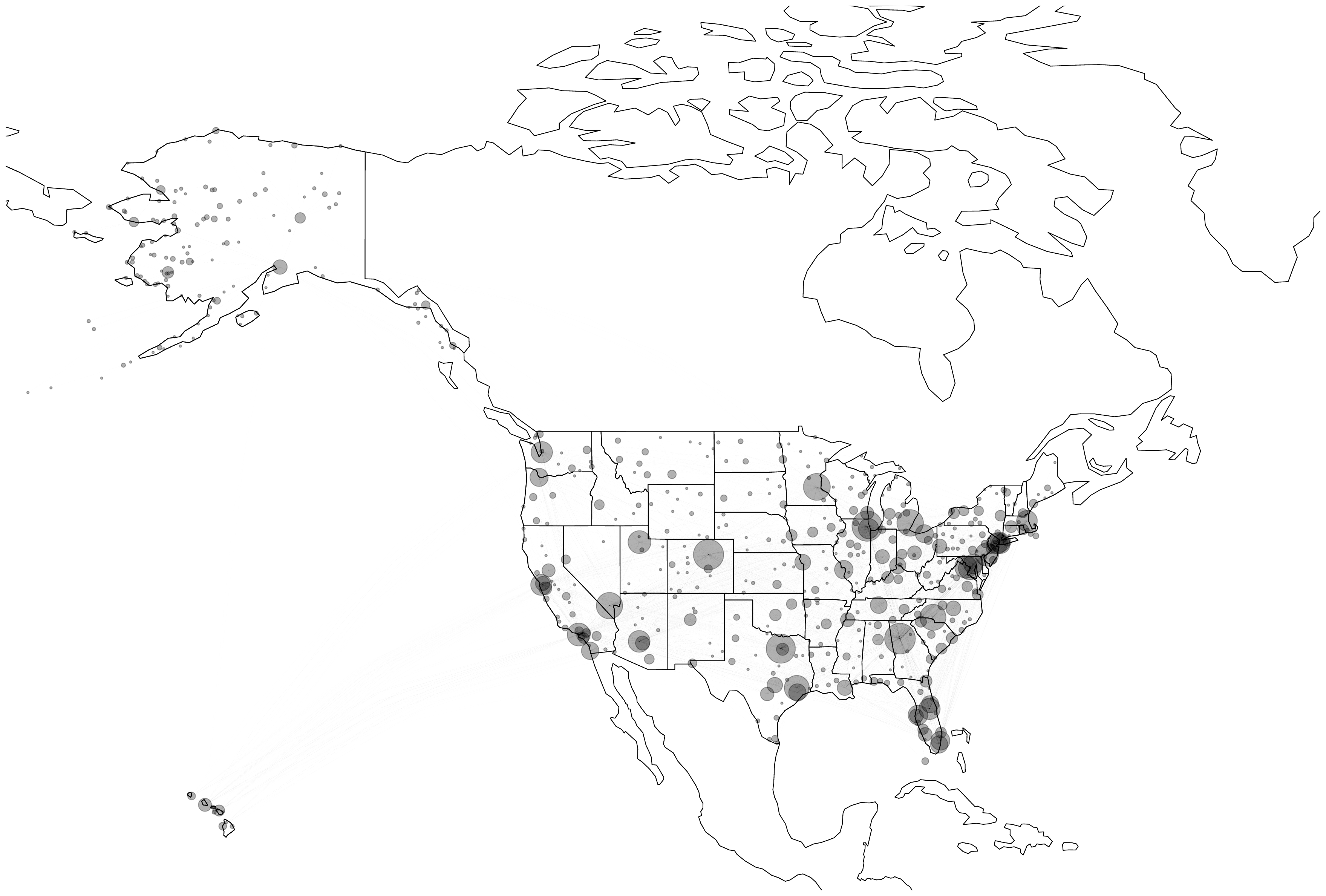}
    \caption{An airport network of the United States of America (size of vertices proportional to their degrees).}
    \label{fig:routes}
\end{figure}

Our goal is to derive a representation that is able to embed how an airport is connected. We assume that there is a hierarchy among airports: international airports act as hubs to smaller, mostly domestic airports. For example, if you wish to fly out of \textit{Santa Fe Regional Airport} (SAF), New Mexico to \textit{Harry Reid International Airport} (LAS), Las Vegas, Nevada, chances are you are going to transit through \textit{Phoenix Sky Harbor International Airport} (PHX), Arizona. Relying only on a visual representation is intricate, and the underlying hierarchy is challenging to highlight, it is thus difficult to distinguish the busiest airports from those having fewer inbound and outbound flights. Furthermore, we wish to encapsulate more than just connectivity between airports, namely the spatial structure of the network and how flights between airports connect states. The question is thus: how can one derive a visual representation of each airport in the network that encompasses its medium haul (domestic) connectivity as well as its local (state-level) neighborhood? 

To that end, let us cluster together the airports of the network based on the state they are located in. By doing so, we obtain fifty groups of airports, each of those corresponding to a U.S. state. Then, by considering the connectivity of each airport to each state instead of one-to-one connection, we can encapsulate local and broader patterns of connectivity. More precisely, we quantify the strength of connectivity of an airport to a state by considering the proportion of airports reached in that state over the total of airports that are served. We thus obtain a visual representation that displays the patterns of connectivity of each airport. The higher the value for a state, the stronger the connection of the airport to the latter.

\begin{figure}[]
    \centering
    \begin{subfigure}{\textwidth}
        \centering
        \includegraphics[width=.6\textwidth]{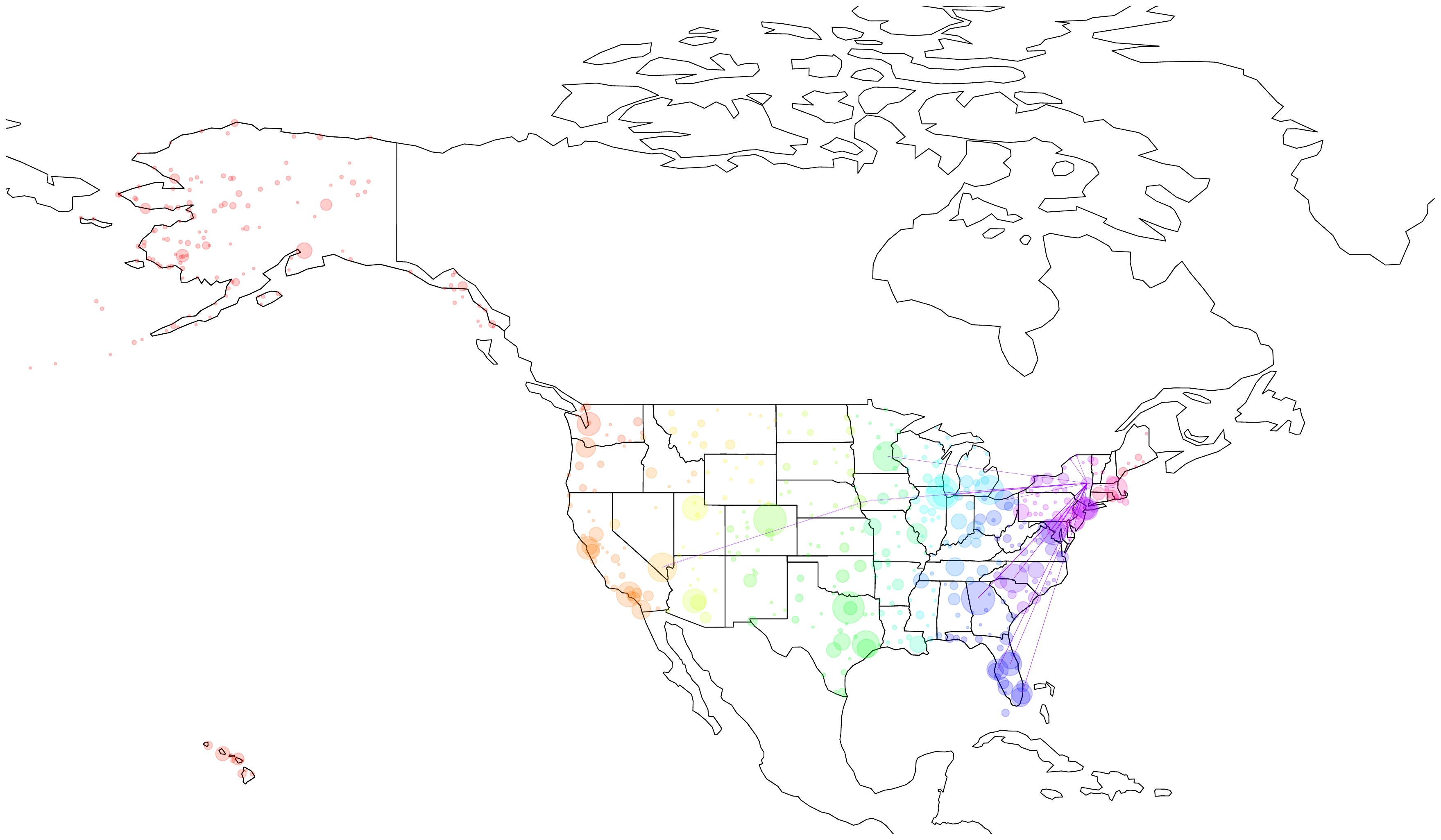}
        \caption{Flights connecting Albany (ALB) to other US airports.}
        \label{subfig:albany_routes}
    \end{subfigure}
    \hfill
    \begin{subfigure}{\textwidth}
        \centering
        \includegraphics[width=\textwidth]{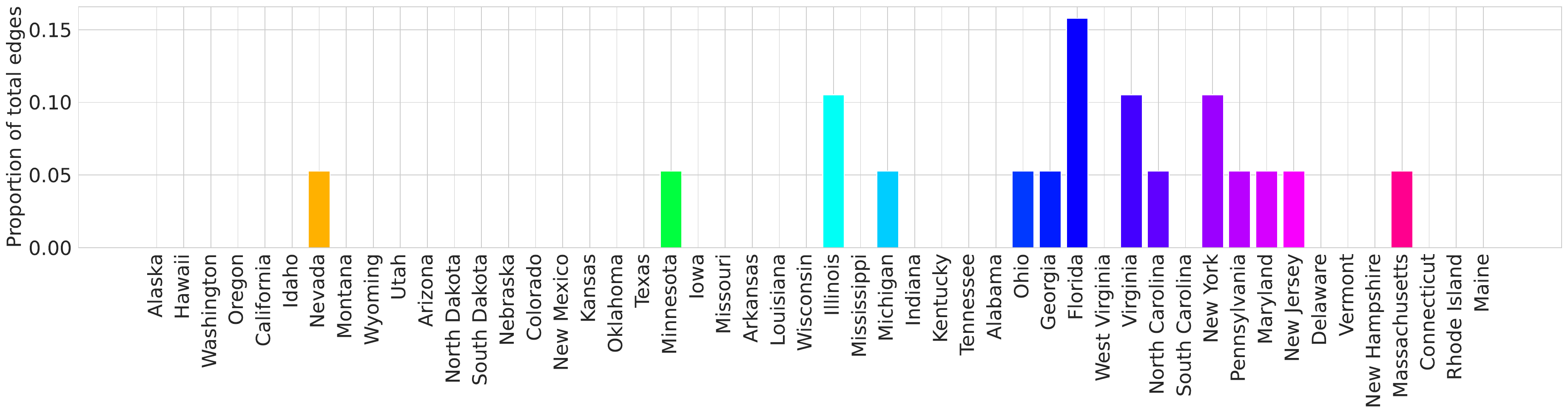}
        \caption{Distribution of the connectivity of Albany International Airport (ALB) to US states.}
        \label{subfig:albany_distr}
    \end{subfigure}
    \hfill
    \ContinuedFloat
    \begin{subfigure}{\textwidth}
        \centering
        \includegraphics[width=.6\textwidth]{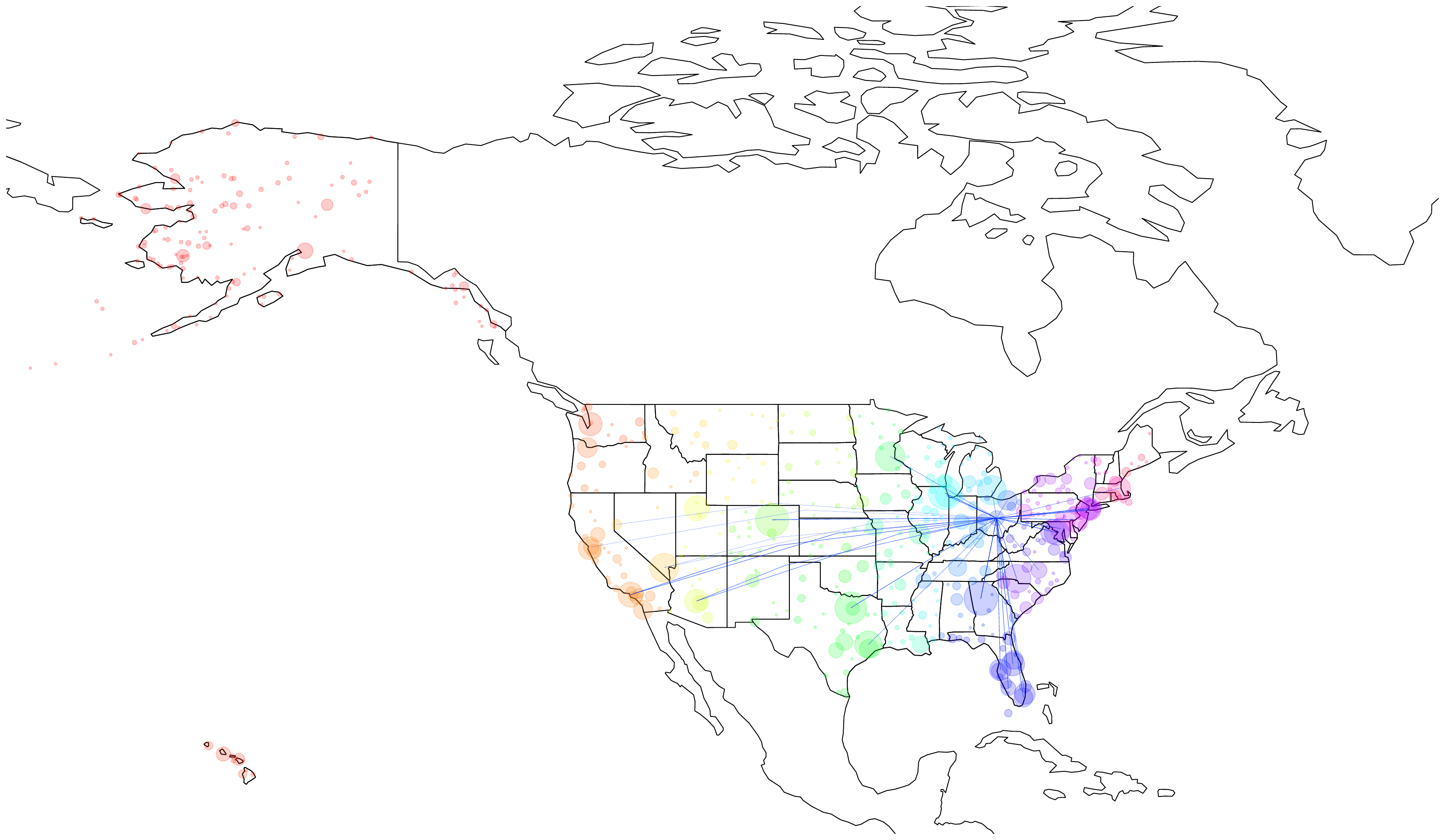}
        \caption{Flights connecting Columbus (CMH) to other US airports.}
        \label{subfig:columbus_routes}
    \end{subfigure}
    \hfill
    \begin{subfigure}{\textwidth}
        \centering
        \includegraphics[width=\textwidth]{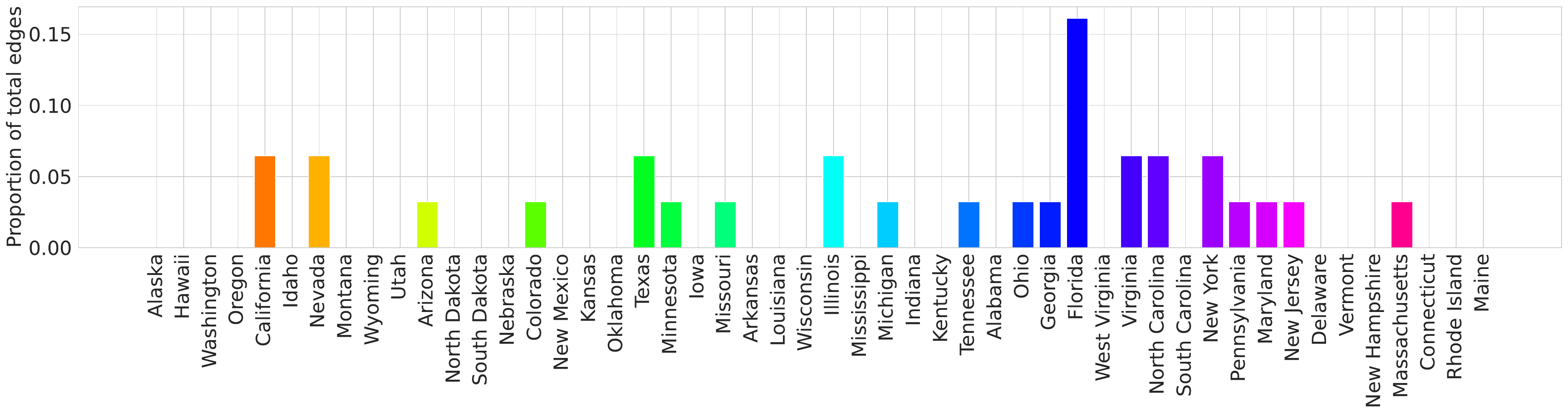}
        \caption{Distribution of the connectivity of Columbus Airport (CMH) to US states.}
        \label{subfig:columbus_distr}
    \end{subfigure}
    \caption{Flights connected to Albany and Columbus and distribution of connections towards states.}
    \label{fig:albany_columbus}
\end{figure}

Let us demonstrate the relevance of this visual representation through two examples. We can first consider two airports on the east coast of the USA: \textit{Albany International Airport} (ALB), NY and \textit{John Glenn International Airport} (CMH) in Columbus, IL. Connections between ALB and CMH are presented in Figure \ref{fig:albany_columbus} \subref{subfig:albany_routes}, \subref{subfig:columbus_routes}. These two airports do not play a major role in their respective states of New York and Illinois but a role of hub for domestic and international flights. The distribution of these airports over the states in Figure \ref{fig:albany_columbus} \subref{subfig:albany_distr}, \subref{subfig:columbus_distr} is similar, mostly to airports in the northeast, midwest and south of the country. On top of the connectivity to each individual state, the color gradient reveals another level of hierarchy related to the different regions of the west, southwest, midwest, southeast and northeast. From the distribution of connectivity in Figure \ref{fig:albany_columbus} \subref{subfig:albany_routes}, \subref{subfig:columbus_routes}, ALB and CMH are primarily connected to airports in the East to Midwest region. 

\paragraph{Extracting vectors.} This pipeline to extract these visual representations is, in fact, a node embedding approach. It can derive node embeddings that are interpretable by design: each component of the vector space constructed is related to a state where airports are located. The representation is a measure of the strength of connection between an airport and each state, thus providing a representation in lower dimension than the one provided by the full network, over a tangible structure—the partition of airports grouped by state. Moreover, vectors are sparse, as not every airport is connected to every state. These embeddings are inexpensive to compute and produce visually interpretable vectors. The question we will attempt to answer is the following: how can such a pipeline be applied to all kinds of networks to produce sparse interpretable embeddings?

The representation of the nodes from the airport network is achieved through the usage of an intermediary grouping of the nodes. By grouping nodes into clusters, the graph could be represented in a bipartite form. In the case of our airport network, the partition of airports according to their respective state produces a bipartite graph with airport nodes and state nodes. Other graphs exhibit a natural bipartite organization: co-authorship networks are naturally bipartite, as authors are connected by papers they co-publish, social networks are also bipartite, friends are connected by their school, family, firm, etc. In our case, the partition of airports according to their respective state is an approximate attempt at projecting the original graph as bipartite. Indeed, projecting back the bipartite graph to a one-mode form would yield edges absent from the original network. Based on this approximation, we detail our node embedding framework in Section~\ref{sec:algorithmic_framework}. Grouping similar nodes in clusters is at the heart of the \LDBGF philosophy of its two implementations we introduce in this paper: \SINrNR and \SINrMF that produces sparse and interpretable node representations.

\section{Algorithmic Framework}
\label{sec:algorithmic_framework}

\subsection{Intuition behind LDBGF}

Embedding techniques aim at providing a lower-dimensional representation of data that encompass structural properties of the units to be represented.
More specifically, graph embedding provides representations of nodes in a graph while retaining topological properties.
We introduced the Lower Dimension Bipartite Graph Framework (\LDBGF), a node embedding framework aimed at producing sparse and interpretable vectors in \emph{Prouteau et al.}~\citep{prouteau_sinr_2021}. 
The intuition behind \LDBGF originates from the observation by \emph{Guillaume and Latapy}~\cite{GL04}  that all graphs can be represented by a bipartite structure.
Some networks lend themselves more naturally to a bipartite representation, like a co-authoring network $G = (\top, \bot, E)$ connecting the set of authors $\bot$ to the set of papers $\top$ they co-signed.
Retrieving the unipartite graph is as easy as projecting the bipartite graph using the $\top$-nodes—adding an edge between any pairs of $\bot$ nodes representing two authors that collaborated.

\begin{figure}[h!]
    \begin{subfigure}[b]{.5\textwidth}
    \resizebox{\textwidth}{!}{
        \input{figures/tikz/ldbgf/full_graph.tikz}
        }
        \caption{A graph $G=(V,E)$ with $8$ vertices and $12$ edges.}
    \end{subfigure}
    \begin{subfigure}[b]{.5\textwidth}
        \resizebox{\textwidth}{!}{
        \input{figures/tikz/ldbgf/full_graph_adjacency.tex}
        }
        \caption{The adjacency matrix $\mathcal{A}$ of $G$. The dimension of the vector for each vertex is $|V|=8$.}
        \label{subfig:adjacency-full-graph}
    \end{subfigure}%
    
    \begin{subfigure}[b]{.5\textwidth}
    \resizebox{\textwidth}{!}{
        \input{figures/tikz/ldbgf/bipartite_cliques.tikz}
        }
        \caption{Projection of $G$ into bipartite graph $G^\prime=(\top,\bot, E^\prime)$ with the minimum number of cliques to cover $E$.}
        \label{subfig:bipartite-cliques}
    \end{subfigure}
    \begin{subfigure}[b]{.5\textwidth}
        \resizebox{\textwidth}{!}{
        \input{figures/tikz/ldbgf/bipartite_adjacency.tex}
        }
        \caption{Adjacency matrix of bipartite graph $G^\prime$. The dimension of the vector for each vertex is the number of cliques (4) used to cover the edges.}
        \label{subfig:adjacency-bipartite}
    \end{subfigure}
    \caption{{Illustration of the \LDBGF, vertices are linked to the cliques they belong to.}}
    \label{fig:LDBGF}
\end{figure}
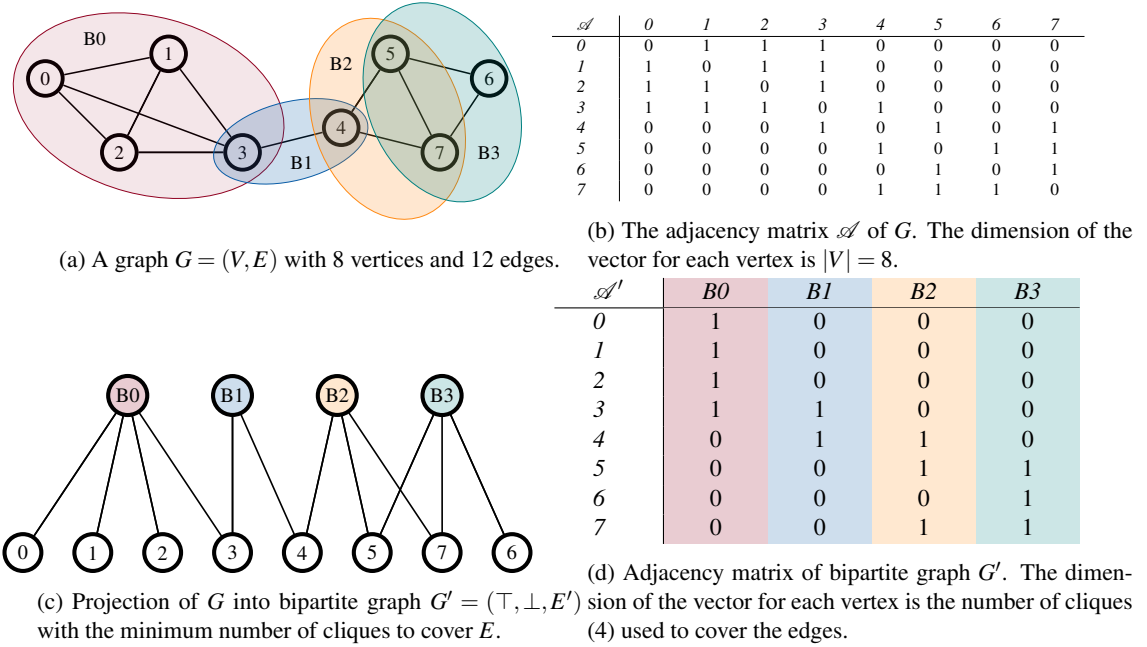

The \LDBGF approach illustrated in Figure \ref{fig:LDBGF} assumes that a latent low-dimensional bipartite structure can be found to represent a graph in a low-dimensional space.
A direct consequence of this representation is a strong intrinsic interpretability of the embedding space produced. Indeed, nodes are represented by their connectivity to the $\top$-nodes from the bipartite structure that are tangible objects.
\textit{Guillaume and Latapy}~\cite{GL04} use cliques in the $\top$-part of the bipartite graph, as presented in Figure \ref{subfig:bipartite-cliques}. As the goal with the \LDBGF is to represent nodes in a lower-dimension space, we enforce the number of $\top$-nodes to be as low as possible. In our example, the adjacency matrix of the bipartite graph $G^\prime$ (\ref{subfig:adjacency-full-graph}) is of lower dimension than that of the original graph $G$ (\ref{subfig:adjacency-bipartite}). Obviously, a dimension reduction may only be observed if the number of cliques is lower than the number of vertices. In that setting, the \LDBGF is related to an \textsc{edge clique covering} problem, namely finding the minimum number of cliques to cover the set of edges~\cite{erdoos1988clique}.
This problem is known to be NP-Hard. Thus we need an alternative to cliques to produce embeddings with low compute requirements.

\subsection{Community detection}

Communities can be considered a relaxation of cliques, so instead of trying to solve an \textsc{edge clique cover} problem, we address a community detection problem. 
Let us consider an (un)directed (un)weighted graph $G=(V,E,\Omega)$, $V$ being the set of vertices, $E$ the set of edges and $\Omega$ the set of weights associated to each edge $(u,v) \in E |  u,v \in V$, $n=|V|$ and $m=|E|$ are respectively the number of vertices and edges in $G$. We define the neighborhood of a vertex $u$ as $N(u)$ and the degree of such a vertex as $d(u)=|N(u)|$, the weighted degree of a vertex $u$ is such that $d_{w}(u)= \sum_{u \in N(v)} \Omega_{u,v}$.
A community structure of $G$ is a partition $\mathcal{C} = \{C_0,\dots,C_j\}, 1\leq j \leq n$ of $V$ into subsets such that the intracommunity density of edges in subgraph $G[C_i]$ is dense and the intercommunity density of edges between $C_i$ and $C_j$ is scarce.

Communities can be uncovered in a myriad of networks, \emph{Girvan and Newman}~\cite{GN02} show for example that social and biological networks boast such community structures. 

To detect communities in this paper, we use the \louvain algorithm introduced by \emph{Blondel et al.}~\cite{blondel_fast_2008}. relying on modularity to detect communities. 
The \louvain algorithm has been extensively used thanks to its quasi-linear time complexity and unsupervised nature without parameters.

Although very efficient, the \louvain algorithm suffers from the resolution limit related to the modularity optimization that is employed. Indeed, the resolution limit is such that communities smaller than a certain size might be difficult to uncover by optimizing modularity and can lead to badly connected communities—communities might be internally disconnected \cite{fortunato_resolution_2007}. To counter the resolution limit, the modularity can be parameterized with multi-resolution parameter $\gamma$. The $\gamma$ parameter \cite{lambiotte2013multi} acts as a bias to increase the weight associated to the probability of the two nodes being connected based solely on their degree. This allows to parameterize the \louvain algorithm and control the granularity of communities.

\subsection{\SINrNR and \SINrMF: community-based approaches to implement \LDBGF}

In \emph{Prouteau et al}.~\cite{prouteau_sinr_2021}, we introduced \SINr as an implementation of \LDBGF that circumvents the complexity of the \textsc{edge clique cover} by supplanting cliques with communities. Community detection, even on large graphs, can be performed very efficiently thanks to the low algorithmic complexity of detection methods. 
In this paper, we introduce implementations of \LDBGF: 
\begin{enumerate}
    \item \SINrNR (Node Recall), an improvement of \SINr that relies on the community structure detected with the \louvain algorithm using multiscale resolution and an ad hoc method to measure the strength of connection between each vertex and each community;
    \item \SINrMF (Matrix Factorization) leverages gradient descent to find the matrix allowing the transition from the graph adjacency matrix to the community membership matrix also extracted with the \louvain algorithm using multiscale resolution. 
\end{enumerate}

We tried several community detection algorithms, as it is central in both these implementations of \LDBGF. Notably, \labelprop~\cite{raghavan_near_2007} (\lp), \infomap~\cite{bohlin_community_2014}, \OSLOM~\cite{lancichinetti_OSLOM_2011}, and \louvain~\cite{blondel_fast_2008}. Although efficient, \lp leads to many small communities for some applications of the framework, leading to vectors of higher dimension than with other algorithms. \infomap is known to perform better but is slower than \louvain and \lp, so is \OSLOM. Although \louvain presents some challenges related to the resolution limit, it performs better than other algorithms we tested on evaluation tasks and is a lot more time-efficient. We ultimately chose to implement these approaches with \louvain. Furthermore, the number of communities can be controlled with the multiscale ($\gamma$) parameter in the modularity measure, and we will see Section~\ref{sec:experiments} that it is important.

\begin{figure}

    \begin{subfigure}[b]{.38\textwidth}{
        \centering
        \resizebox{\textwidth}{!}{
            \input{figures/tikz/sinr/full_graph_communities.tikz}
        }
        }%
        \caption{A graph $G=(V,E)$ partitioned in two communities.\\} 
        \label{subfig:communities}
    \end{subfigure}
    \begin{subfigure}[b]{.62\textwidth}{
        \resizebox{\textwidth}{!}{
            \input{figures/tikz/sinr/bipartite_communities.tikz}
            }
            }
            \caption{Bipartite projection of $G$ into graph $G^\prime = (\top, \bot, E^\prime)$ along the communities. Weight on the edges is the \emph{NR} value regarding the community—the proportion of neighbors in that community. }
            \label{subfig:bipartite-SINr}
    \end{subfigure}
    \begin{subfigure}[b]{\textwidth}
    \centering
    \input{figures/tikz/sinr/sinr_emb_matrix.tex}
    \caption{Adjacency matrix of $G^\prime$, each row is a \SINrNR embedding.}
    \label{subfig:SINr-emb}
    \end{subfigure}

    \caption{Illustration of \SINrNR, vertices are represented based on the communities they are linked to.}
    \label{fig:SINr}
\end{figure}
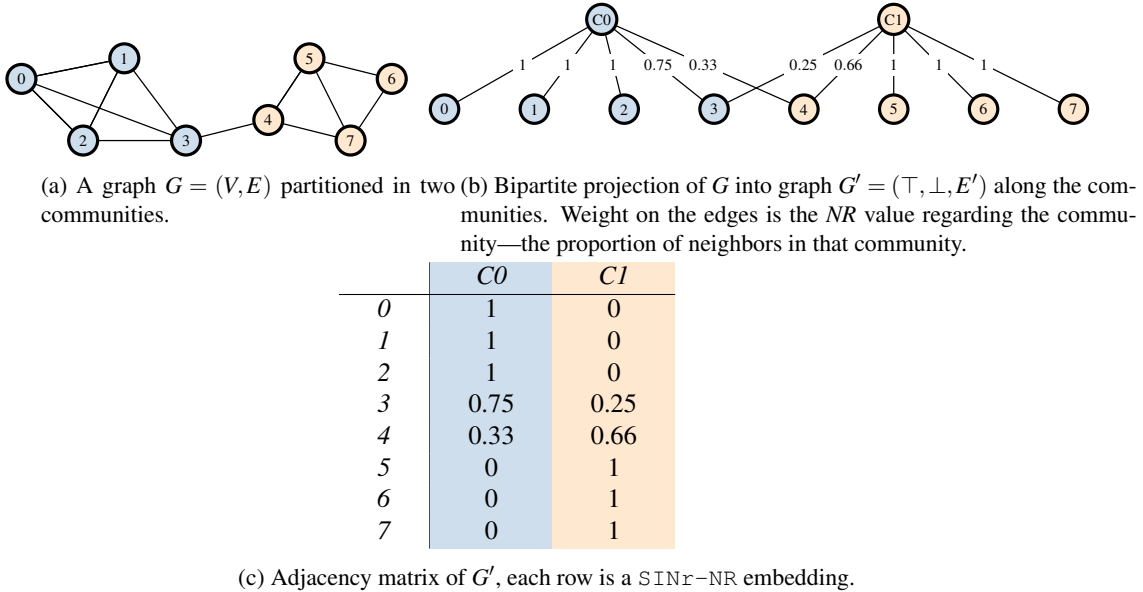

\paragraph{\SINrNR.} To derive embeddings from a graph with \SINrNR as illustrated in Figure~\ref{fig:SINr}, one first need to detect its communities. Applying the \louvain method to a graph $G$ produces a partition of the graph's vertices (\ref{subfig:communities}) which is used to build an embedding vector for each vertex. Based on the partition in communities, \SINrNR weights the vector using the node recall (NR) measure introduced in \emph{Dugué et al.}~\cite{dugue_bringing_2019} that is equivalent to \emph{Lancichinetti et al.}'s \emph{embeddedness}~\cite{lancichinetti_characterize_2010} of a vertex in its community. With \SINrNR, we are essentially quantifying the distribution of weighted degree of a vertex $u$ towards each community $C_i$ (\ref{subfig:bipartite-SINr}). Given a vertex $u$, a partition of the vertices $\mathcal{C} = \{C_0,\dots,C_j\}$, and $C_{i}$ the $i_{th}$ community so that $1 \leq i \leq j$, the node recall of $u$ considering the $i^{\text{th}}$ community is :

\begin{equation}
    \text{NR}_{i}(u) = \frac{d_{C_{i}}(u)}{d(u)} \text{ with } d_{C_{i}}(u) = \sum_{v \in C_i} \Omega_{uv}
\end{equation}

Upon computing \emph{NR} for one vertex over all the communities, one obtains a vector representing the vertex considered. When computed for all vertices in the graph, we obtain the embedding model. The embedding vectors are thus sparse (\ref{subfig:SINr-emb}), since not every vertex is connected to a node in every community previously uncovered, their dimension is linear with the number of communities.

\paragraph{\SINrMF.} 
The second implementation of \LDBGF we introduce in this paper does not rely on an ad hoc measure of strength of connection such as \emph{NR}. We named it \SINrMF for Matrix Factorization: it attempts to factorize the adjacency matrix $\mathcal{A}$ of a graph $G$ into the product of two matrices $U$ and $C$, the community membership matrix extracted using \louvain. In our case, the matrices $\mathcal{A}$ and $C$ are known. By letting gradient descent handle the optimization of $U$ we hope to avoid the ad hoc \emph{NR} measure of \SINrNR and obtain $\mathcal{A}$, an approximation of the graph at hand in the space described by $C$. The goal of \SINrMF is thus to minimize the difference between $\mathcal{A}$ and the product of $U$ and $C$ in the following model:

\begin{equation}
    \text{SINr-MF}(G) = \argmin\limits_{U}(\text{MSE}(\mathcal{A}, UC^T)) 
\end{equation}
This optimization is performed by gradient descent using a \emph{Mean Squared Error} (MSE) loss.

\paragraph{}The code for these implementations is available on GitHub\footnote{\url{https://github.com/SINr-Embeddings}} and has been implemented using Python and the efficient \emph{Networkit} library of \emph{Staudt et al.}~\cite{StaudtSM14}.

The next section evaluates thoroughly the ability of \SINrNR and \SINrMF at embedding graph properties at different levels. To do so, we consider multiple tasks, \textit{including link prediction}, \textit{vertex degree regression} or \textit{label prediction}. We also deal with \nlp tasks and evaluate the interpretability of our approach on textual data.

\section{Experiments}
\label{sec:experiments}

Properties of complex networks may be analyzed at different levels. We distinguish three: microscopic, mesoscopic and macroscopic. These levels correspond to the scale at which we examine the network, from local interaction of individual vertices (microscopic) to larger structures within the network such as communities or motifs (mesoscopic), to the network as a whole (macroscopic).

Embeddings are meant to project data in a vector space that encompasses their relations and organization. Subsequently, well-formed graph embeddings should carry over structural characteristics from the original graph. We evaluate the propensity of graph embedding methods to encapsulate key topological information from the network at these three levels in the following sections. We first consider the microscopic level, the goal is to assess the capacity of vectors to model local interactions between vertices with experiments such as \emph{link prediction}, \textit{vertex degree regression} or \textit{vertex clustering coefficient regression}. At the mesoscopic level, our experiments are targeted towards intermediate structures such as communities and include: \emph{vertex clustering}, \emph{vertex classification}. Regarding the macroscopic level, which relates to the graph as a whole, we evaluate the capacity of embedding to model \pagerank through \emph{PageRank score regression}.

{We then focus on the specific application of our method to word co-occurrence network to derive word embeddings. We demonstrate that our approach is relevant in this context through two evaluations: \textit{pairwise word similarity}, and \textit{word categorization}. Furthermore, we study the stability aspect of our algorithms when it comes to word embeddings.}

Following these experiments aimed at measuring the capacity of graph embedding methods to grasp graph topology, we also evaluate our approach on \textit{interpretability} of words embeddings from the vectors space. Finally, most embedding methods are notoriously non-deterministic when it comes to producing vectors. Since it is a desirable feature in interpretable settings, we also evaluate the \textit{stability} of \SINrNR against \FWtoV and \FSPINE.

We chose five real-world graphs composite in fields and sizes ($n=|V|$ and $m=|E|$) for our experiments. The variety in network sizes allows studying the scalability of our approach regarding the downstream evaluation tasks. For the sake of our experiments, we consider each graph as undirected and extract their largest connected component.
\begin{itemize}
    \item[a.] \FCTS (\CTS; $n=2,110$; $m=3,720$) and \COR ($n=2,485$; $m=5,069$) are networks of scientific citations.
    \item[b.] \FEML (\EML; $n=986$; $m=16,687$) is a sender-recipient email network within an European research institution.
    \item[c.] \ARX ($n=17,903$; $m=196,972$) is a co-authorship network of articles published on ArXiv in the Astrophysics category between January 1993 and April 2003.
    \item[d.] \FFB (\FB; $n=63,392$; $m=816,831$) is a graph of user friendships from Facebook.
\end{itemize}

 To assess \SINrNR and \SINrMF we also introduce state-of-the-art methods if applicable to the evaluation task being performed. {Our main source for graph embedding models was the \texttt{karateclub} library implemented in Python. However, it did not include Graph Neural Network approaches that are state-of-the-art for some graph-related tasks.} We selected \FDWLK (\DWLK) and \FWKLT (\WKLT) that are based on random walks. \HOPE factorizes similarity matrices. \FVERSE relies on a single layer neural network to preserve similarity between vertices. \FLouvainNE \cite{bhowmick_louvainne_2020} relies on hierarchical community detection to derive node embeddings. The implementations for \FDWLK, \FWKLT and \HOPE are provided by the \texttt{karateclub} library implemented in Python. \FLouvainNE and \FVERSE are the implementations provided by the authors respectively in C and C++. The default parameters of the implementations are kept for each node embedding baseline, nodes are embedded in 128 dimensions. For \SINrMF, the number of epochs is set to 3000 with stochastic gradient descent and a learning rate of $5e^{-3}$. For \SINrNR and \SINrMF, the $\gamma$ parameter that impacts the number of communities is discussed in the results in Section \ref{subsec:exp_microscopic_level}.

\subsection{{Run-time}}

Before proceeding with experiments at the three levels previously introduced, let us consider the compute time required for each method. Methods with a low run-time are more energy efficient and also valuable for end users since they do not need to wait for lengthy computation before exploiting a vector space. Furthermore, as discussed in Section~\ref{sec:related_work}, we aim to design an approach with low environmental impact, the literature having demonstrated the high CO$_2$ emissions of recent large architectures. We thus compute the wall time and CPU time required by each model on each of our baseline over ten runs and present the averaged results in Table~\ref{tab:runtime}. 

\begin{table}[h!]
\centering
\resizebox{\textwidth}{!}{%
\begin{tabular}{l|cc||ccc||cc}
\hline
 &
  \SINrNR &
  \SINrMF &
  \DWLK &
  \WKLT &
  \HOPE &
  \FLouvainNE &
  \FVERSE \\ \hline
\COR & \textbf{0.04/0.04} & 8/161 & 10/36 & 16/31 & 0.22/9 & 0.10/0.09 & 83/331 \\
\EML & 0.11/0.11 & 3/50 & 5/13 & 8/14 & 0.55/20 & \textbf{0.09/0.08} & 34/138 \\
\CTS & \textbf{0.03/0.03}  & 6/123 & 8/22 & 14/25 & 0.19/7 & 0.05/0.04 & 69/274 \\
\ARX & 1.5/1.5 & 1.2K/20K & 289/580 & 435/812 & 25/600 & \textbf{0.79/0.72} & 610/2.4K \\
\FB  & 6.7/6.7  & 6.5K/116K & 414/822 & 646/1.2K & 61/638 & \textbf{3.1/2.8} & 2.8K/11K \\ \hline
\end{tabular}
}
\caption{Average run-time (left) and total CPU time (right, with parallelism) in seconds over 10 runs. Run-time is computed with two \emph{Intel Xeon E5-2690 v2 3.00GHz CPU} : 20 cores, 250Go RAM.}
\label{tab:runtime}
\end{table}

\SINrNR's run-time is the lowest among graph embedding methods implemented in Python. Even on larger graphs, its run-time is inferior to its counterparts by a few orders of magnitude. When we take into account \FLouvainNE which is implemented in C, even on the largest graph, \FFB (\FB), \SINrNR is only a few seconds behind. \HOPE is the second best algorithm in terms of run-time in the Python category. \SINrMF on the other hand, is relatively efficient on smaller graphs. However, on larger graphs, it reaches a point where it is not sustainable compared to other methods. It is, however, important to note that \SINrMF was purposefully trained using CPUs instead of GPUs, which are far more efficient at processing matrices. Still, because of its run-time, we do not consider \SINrMF in the rest of this paper when dealing with \nlp because of the size of textual graphs. We still thoroughly evaluate the approach on all the graphs tasks.

\subsection{{Assessment of the quality of vertex embedding}}

\subsubsection{Microscopic-Level}
\label{subsec:exp_microscopic_level}

\paragraph{Link Prediction.} Let $G=(V,E)$ be an undirected graph, $U$ the universal set containing $\frac{n(n-1)}{2}$ possible edges in $V$ and $E^\complement = U \setminus E$. The link prediction task is set up as a binary classification of edges into two classes, either existing or absent. A classifier is trained to detect whether an edge is likely to exist or appear in the graph at hand.

As in \cite{grover_node2vec_2016, tsitsulin_verse_2018, makarov_survey_2021}, we randomly select pairs of vertices at the extremities of edges from the graph (20\%) into a test set. Subsequently, these edges are removed from the graph under the constraint of not increasing the number of components. The remaining edges in $G$ are part of the training set. Embedding vectors for each model are trained using the train set only. The set of negative examples for the link prediction is sampled in $E^\complement$ with similar proportions to the positive train and test sets. An \XGB classifier is then trained to discriminate between existing and non-existing edges for a given pair of nodes. The representation used in input of the classifier for an edge $(u,v)$ is the Hadamard product between the two embedding vectors of vertices $u$ and $v$ in each model. The task is performed 50 times for each network and each model, and the average accuracy is presented in Table~\ref{tab:linkprediction}. For this task, we also consider \FHTS (\HTS), a feature engineered method relying on heuristics features that was shown to be very efficient. As in Sinha et al.~\cite{sinha2019systematic}, the features involved are the following : \textit{Common Neighbors}, \textit{Adamic Adar}, \textit{Preferential Attachment}, \textit{Jaccard Index} and \textit{Resource Allocation Index}.

\begin{table}[h!]
\centering
\resizebox{\textwidth}{!}{%
\begin{tabular}{@{}l|cc||cccccc@{}}
\toprule
                                         & \SINrNR & \SINrMF     & \HTS & \DWLK & \WKLT & \HOPE & \FLouvainNE & \FVERSE \\ \midrule
\COR & 0.845          & \textbf{0.850} & 0.728          & 0.708    & 0.773    & 0.760 & 0.827     & 0.809 \\
\EML & 0.860 & 0.798          & \textbf{0.885}          & 0.790    & 0.876    & 0.847 & 0.752     & 0.852 \\
\CTS & 0.877          & \textbf{0.879} & 0.755          & 0.736    & 0.820    & 0.832 & 0.863     & 0.859 \\
\ARX & 0.930          & 0.893          & \textbf{0.980} & 0.912    & 0.954    & 0.914 & 0.847     & 0.957 \\
\FB  & 0.915          &  0.892             & \textbf{0.930}          & 0.859    & 0.920    & 0.900 & 0.847    &  0.917     \\ \bottomrule
\end{tabular}%
}
\caption{Average accuracy for the link prediction task over 50 runs.}
\label{tab:linkprediction}
\end{table}

Overall, both \SINrNR \& \SINrMF are on par with competing methods, especially on small networks. \SINrNR is consistently better than \HOPE and \FDWLK (\DWLK) across all networks and close to \FWKLT (\WKLT). The matrix factorization approach \SINrMF is leading on the two smallest networks (\COR \& \CTS) closely followed by \SINrNR. It seems that \SINrMF is less efficient when networks become larger, but \SINrNR remains competitive with the leading method \FHTS (\HTS) even on networks of higher magnitude.

\begin{figure}[h!]
    \centering
    \begin{subfigure}[t]{0.49\textwidth}
        \centering
        \includegraphics[width=\textwidth]{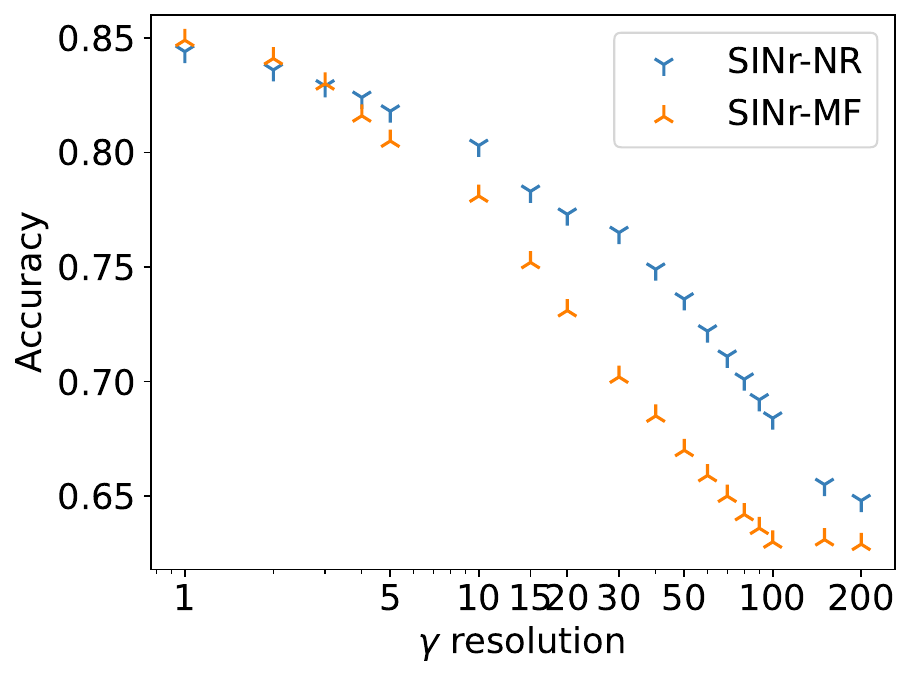}
        \caption{Cora}
        \label{subfig:cora}
    \end{subfigure}%
    ~ 
    \begin{subfigure}[t]{0.49\textwidth}
        \centering
        \includegraphics[width=\textwidth]{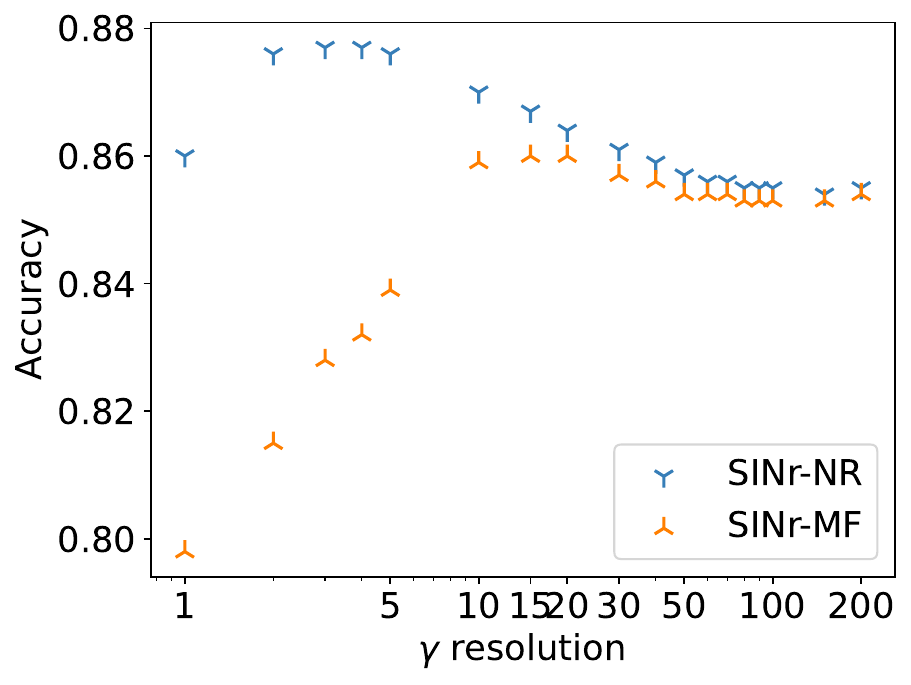}
        \caption{Email-EU}
        \label{subfig:email}
    \end{subfigure}
    ~
    \begin{subfigure}[b]{0.49\textwidth}
        \centering
        \includegraphics[width=\textwidth]{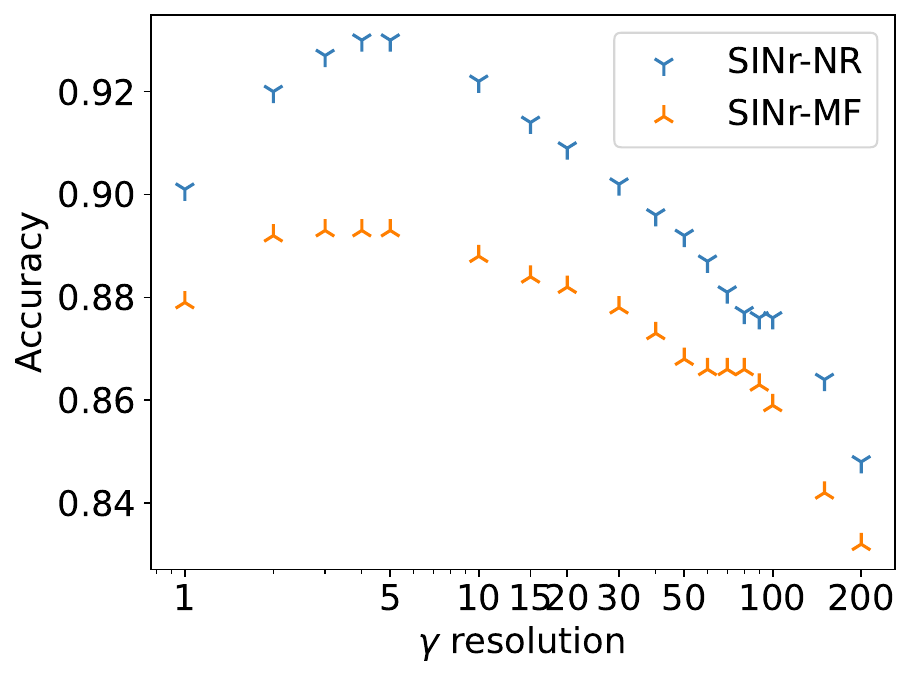}
        \caption{arXiv}
        \label{subfig:arxiv}
    \end{subfigure}
    \begin{subfigure}[b]{0.49\textwidth}
        \includegraphics[width=\textwidth]{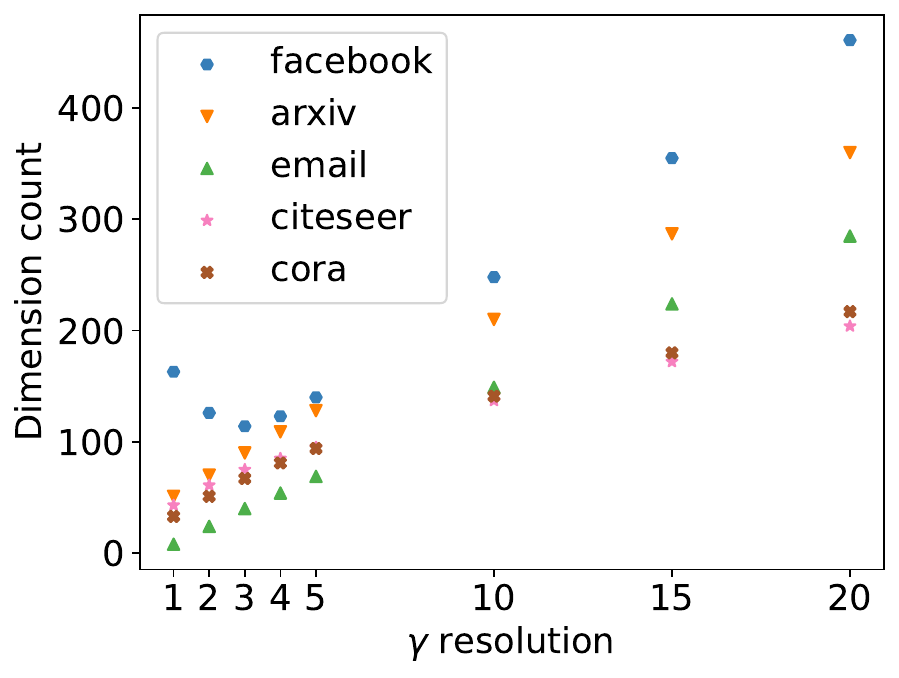}
        \caption{Average number of dimensions according to $\gamma$.}
        \label{subfig:dimensions}
    \end{subfigure}
    \caption{Average accuracy (over 50 runs) on Link Prediction according to $\gamma$ with \SINrNR in yellow and \SINrMF in blue for Cora, Email-EU and ArXiv. Average number of dimensions of our models (over 50 runs) according to the $\gamma$ is given in (d).}
    \label{fig:lp-gamma}
\end{figure}

The only parameter of our methods is the $\gamma$ multi-resolution parameter inherited from \louvain, which allows controlling the dimension of vertex embedding vectors. Higher values of $\gamma$ result in vectors of higher dimension. Depending on the size of the network at hand, it might be beneficial to have more dimensions in the embedding space to better represent the complexity of the network structure. To that end, we plot Figure~\ref{fig:lp-gamma} the accuracy on the link prediction task according to the $\gamma$ value for \louvain. For the smallest network \COR, the maximum accuracy is reached for a $\gamma$ value of 1. Our intermediary \EML network admits a maximum accuracy when $\gamma$ is $3$ for \SINrNR when \SINrMF needs more dimensions to reach its highest accuracy on link prediction, setting the $\gamma$ between 10 and 20. On \ARX, the best accuracy is reached with a $\gamma$ set to values around 5. A good rule of thumb is that the larger the graph is, the more dimensions are needed to represent vertices' interactions and thus a higher $\gamma$ value needs to be chosen. As a consequence of these results, the $\gamma$ value chosen for \COR, \FCTS and \EML is 1, when $\gamma$ is set to 5 for \ARX and \FFB for all the experiments on these networks. We also consider the number of dimensions of our models according to $\gamma$. As a direct consequence of its definition, increasing gamma mechanically increases the number of communities, and thus number of dimensions. However, as one can see in Figure~\ref{subfig:dimensions}, when $\gamma$ is set to $1$, for the small networks, the number of dimensions is lower than $128$, it is actually $8$ for Email-EU, $33$ for Cora and $43$ for Citeseer. With $5$ for arXiv and Facebook, it is close to $128$ (respectively $128$ for arXiv and $140$ for Facebook), the number of dimensions chosen for the competing approaches.

\paragraph{Vertex degree regression.}  In order to evaluate the capacity of embedding models to encapsulate information local to a vertex, we proceed to predict the degree of vertices in the graph. From an undirected graph $G$, we first learn vertex embeddings. Then, a linear regression is trained from the embedding, the target value for each vertex is extracted from the degree sequence of $G$. The set of vertices is split randomly, with 80\% of the vertices in the training set and the remaining 20\% in the test set. Regression is performed 50 times for each model and the average R-squared ($\text{R}^2$) coefficient of determination is presented in Table \ref{tab:degree_regression} to measure the goodness of fit.

\begin{table}[h!]
\centering
\resizebox{\textwidth}{!}{%
\begin{tabular}{@{}l|cc||ccccc@{}}
\toprule
         & \SINrNR                            & \SINrMF & \HOPE  & \FDWLK & \FWKLT & \FLouvainNE & \FVERSE \\ \midrule
\COR     & \textbf{1.000} & -0.586 & 0.773 & -0.122   & -0.028   & -0.067    & 0.227 \\
\CTS & \textbf{0.998} & -0.077 & 0.741 & 0.140    & 0.342    & 0.037     & 0.010 \\
\EML    & \textbf{1.000} & 0.022  & 0.990 & 0.388    & -0.783   & -5.183    & 0.869 \\
\ARX    & \textbf{1.000} & 0.543  & 0.937 & 0.419    & 0.725    & 0.144     & 0.585 \\
\FB & \textbf{1.000} & 0.647  & 0.935 & 0.358    & 0.766    & 0.091     & 0.706 \\ \bottomrule
\end{tabular}
}
\caption{Average (50 runs) $\text{R}^2$ for vertex degree regression.}
\label{tab:degree_regression}
\end{table}

Vertex degree regression results presented in Table~\ref{tab:degree_regression} put \SINrNR ahead of all models. \HOPE is second and \FVERSE third, with varying performances across networks. \SINrMF achieves subpar performances on \COR, \FCTS and \FEML but is in keeping with \FVERSE on \ARX and \FFB. Random walk based methods seem to underperform at embedding degree in their representations on small networks. \FLouvainNE embeddings are unable to help with the prediction of vertex degrees. 

For the datasets considered, performances of \SINrNR on the degree regression may be the direct consequence of the ad hoc embedding weighting method employed. Measuring the relative distribution of weighted degree of each vertex over the network community structure is an indirect encoding of a vertex degree. The more values in the embedding are diffuse, the higher the potential degree of the vertex is, since it is connected to many vertices in varying communities. On the other hand, it appears that \SINrMF matrix factorization approach is not efficient at embedding vertex degrees. \SINrMF is less prone to embedding degree information, since it aims at finding the transition matrix between the network's adjacency matrix and the community membership matrix.

\paragraph{Clustering coefficient regression.} Another aspect of local organization is the connectivity among neighbors of a vertex, measured by the clustering coefficient. 

Similarly to the degree regression experiment previously introduced, we compute the clustering coefficient of each vertex in the graph and learn a linear regression model from the vector representation of each graph and each model. The results presented in Table~\ref{tab:clustering_regression} are the average R$^2$ determination coefficient over 50 runs with a random 80\% train 20\% test split.

\begin{table}[h!]
\centering
\resizebox{\textwidth}{!}{%
\begin{tabular}{@{}l|cc||ccccc@{}}
\toprule
         & \SINrNR  & \SINrMF & \HOPE   & \FDWLK & \FWKLT & \FLouvainNE & \FVERSE  \\ \midrule
\COR     & 0.007 & 0.027  & -0.161 & 0.0001    & -0.020   & \textbf{0.052}     & -0.001 \\
\CTS & \textbf{0.003} & -0.411 & -0.493 & -0.015   & -0.006   & -0.950    & 0.001  \\
\EML    & 0.060 & -0.011 & -0.012 & \textbf{0.065}    & -1.756   & -35.379   & -0.078 \\
\ARX    & 0.247 & 0.236  & 0.126  & 0.325    & \textbf{0.344}    & 0.036     & 0.260  \\
\FB & 0.027 & 0.046  & 0.017  & 0.132    & \textbf{0.167}    & 0.009     & 0.036  \\ \bottomrule
\end{tabular}%
}
\caption{Clustering coefficient regression, average $\text{R}^2$ over 50 runs.}
\label{tab:clustering_regression}
\end{table}

On average, the determination coefficient scores for the clustering coefficient prediction are low. \SINrNR is the sole model to always present a positive R$^2$ coefficient regardless of the network. It seems it is difficult to come with a hard conclusion on small networks (\COR, \CTS and \EML) where R$^2$ values are really low and performance is variable from one network to the other. On larger networks, random walk based methods obtain more encouraging results, followed by \FVERSE, and the two implementations, \SINrNR and \SINrMF. None of the models manage to emerge as a clear contender for vertex clustering coefficient prediction. 

From these observations, one can only wonder why the clustering coefficient cannot be easily predicted from the vertex embedding when degree is a property well embedded in their representation. Clustering coefficient is a measure of the connection among vertex neighbors and is thus rather local to a vertex. However, this interaction is happening at an edge away from the node considered and might be harder to model in the representation. Random walk based models might capture more of this structure on larger graphs with many edges. 

\paragraph{Microscopic-level modeling.} Micro-scale performances of \SINrNR on \emph{link prediction} and \emph{vertex degree prediction} show that it is efficient at embedding local information. In the case of \SINrMF, performances are more varied over our three micro-scale tasks. Although \SINrMF performs satisfactorily on \emph{link prediction}, its lack of performances on \emph{vertex degree regression} and \emph{vertex clustering coefficient regression} outlines its limitations at embedding local information from the vertices.

Moving on from microscopic-level information, we now want to take a step back and evaluate the capacity with which \SINrNR and \SINrMF approaches are able to carry mesoscopic properties.

\subsubsection{Mesoscopic-Level}

\paragraph{}In this section, to evaluate the ability of our approaches to embed mesoscopic information, we cluster and classify vertices into communities in graphs.

\label{subsec:exp_mesoscopic_level}

 Communities are intermediate structures between the vertex and the graph as a whole, making them interesting objects to study the capacity of a model to embed mesoscopic-level information. In our first experiment, we proceeded in an unsupervised manner before adopting a supervised approach in the following experiment. { There is scrutiny around the evaluation of community detection and trying to predict so called \textit{``ground-truth''} communities. The partitions of nodes we use are based on metadata associated with each node. Yet, metadata-based communities may not match the communities detected by a community detection algorithm }\cite{dao_community_2017,chakraborty_metadata_2018}{. This may be the consequence of diverse factors: metadata is not relevant to the network structure, metadata and uncovered partition may capture different aspects of topology, or the network may not exhibit a community structure }\cite{lee2014community,peel2017ground}{. However, in our case, we do not evaluate community detection but rather the ability to find a clustering of nodes in groups. This task is performed with a clustering algorithm applied to the vectors and a classifier. Thus, we can perform our experiment, with the caveat that partitions based on metadata may not correspond to partitions that would be detected by community detection algorithms. A second argument could be made against such an evaluation when working with community-based embeddings. However, although }\SINrNR {and} \SINrMF {work with the community structure to derive a representation, they weight the relations to the latter and do not retain the entirety of community membership.}

 \paragraph{Vertex community clustering.}\COR, \FCTS and \FEML have known ground-truth community structures that allow studying whether communities are encoded in embedding vectors. Following the spectral clustering method \cite{shi_normalized_2000}, we first construct an affinity matrix based on cosine similarity between all pairs of vertices. Using the affinity matrix alleviates the discrepancies in dimensions across embedding methods since the clustering is performed on a $V\times V$ matrix. The average \emph{Normalized Mutual Information} (NMI) \cite{lancichinetti_detecting_2009} score over 50 runs for each network with a known community structure and each method is presented in Table~\ref{tab:spectral_clustering}.

\begin{table}[h!]
\centering
\resizebox{\textwidth}{!}{%
\begin{tabular}{@{}l|cc||ccccc@{}}
\toprule
         & \SINrNR & \SINrMF & \HOPE  & \FDWLK       & \FWKLT & \FLouvainNE & \FVERSE          \\ \midrule
\COR     & 0.364  & 0.047  & 0.368 & 0.020          & 0.296    & 0.311     & \textbf{0.435} \\
\CTS & 0.331  & 0.112  & 0.322 & 0.009          & 0.177    & 0.147     & \textbf{0.406} \\
\EML    & 0.567  & 0.266  & 0.682 & \textbf{0.703} & 0.670    & 0.644     & 0.698          \\ \bottomrule
\end{tabular}
}
\caption{Spectral clustering results based on embedding similarity. Average \emph{NMI} over 50 runs between ground truth labels and predicted clusters.}
\label{tab:spectral_clustering}
\end{table}

Regarding unsupervised detection of communities, scores for \SINrNR are close to that of \HOPE. \SINrMF does not allow to cluster the embeddings efficiently into communities from the similarity between embeddings. Contrary to \FDWLK, all models seem not to show great discrepancies across networks even though it is the best performing model on \FEML. \FVERSE shows abilities to learn useful vectors for community clustering. Although \SINrNR has low results on \FEML, it manages to be respectively third and second-best model on \COR and \FCTS. The major lesson drawn from these scores is that detecting communities from the similarity between embedding is a complex task. Thankfully, we can also adopt a classification approach to the detection of communities.

\paragraph{Vertex community classification.} Community prediction can also be performed in a supervised manner with the help of a classifier. In this setting, we choose the \XGB classifier previously used for the \emph{Link Prediction} task (Section \ref{subsec:exp_microscopic_level}). The classifier is trained using 80\% of the vertices' embeddings and their respective community labels, the remaining 20\% of vertices is used as a test set. The experiment is run 50 times, and we present in Table~\ref{tab:vertex_classification} the averaged accuracy. 

\begin{table}[h!]
\centering
\resizebox{\textwidth}{!}{%
\begin{tabular}{@{}l|cc||ccccc@{}}
\toprule
         & \SINrNR & \SINrMF & \HOPE  & \FDWLK & \FWKLT & \FLouvainNE & \FVERSE \\ \midrule
\COR     & 0.752/0.786  & 0.716/0.786  & 0.799 & 0.719    & \textbf{0.852}    & 0.831     & 0.850 \\
\CTS & 0.690/0.718  & 0.685/0.694  & 0.718 & 0.630    & \textbf{0.758}    & 0.722     & \textbf{0.758} \\
\EML    & 0.432/0.719  & 0.222/0.604  & \textbf{0.721} & 0.686    & 0.700    & 0.672     & 0.656 \\ \bottomrule
\end{tabular}
}
\caption{Vertex classification results, average accuracy over 50 runs. For \SINrNR and \SINrMF, the first number corresponds to $\gamma=1$ when the second is when $\gamma=5$.}
\label{tab:vertex_classification}
\end{table}

Regarding supervised node community labeling, \FWKLT is the best performing model. On smaller networks (\COR and \FCTS), \SINrNR is ahead of \FDWLK and \SINrMF has similar performances. More globally, \SINrNR, \HOPE and \FLouvainNE have performances within a range of eight points and can be said to perform equally. However, a more significant gap between the baseline methods and \SINr's two approaches appears on \FEML (\EML). These results would indicate that \SINrNR doesn't manage to grasp community membership as well as other methods on a network with far more edges and fewer vertices than \COR and \FCTS. 

However, when we slightly increase the $\gamma$ value for community detection on this task, we obtain more dimensions in the representation space and results improve. When we set $\gamma = 5$, \SINrNR and \SINrMF results on \COR respectively jump to $0.786$ when they increase to $0.718$ and $0.694$ on \FCTS. The strongest increase is observed for \EML where changing to a $\gamma$ value of $5$ leads to respective accuracy results of $0.719$ and $0.604$ for \SINrNR and \SINrMF thus making \SINrNR the second-best model. The $\gamma$ parameter has an undeniable impact on performance for \textit{vertex community classification} as for \textit{link prediction} and illustrated in Figure~\ref{fig:lp-gamma}.

\paragraph{Mesoscopic-level modeling.} Our evaluation of mesoscopic-level information in graph and word embeddings shows that models are capable of embedding more distant structures and context than just the vertex and its neighborhood. Although clustering vertices from the similarity between vectors is a complex task on which most models do not perform, classifying vertices into communities from their embeddings is possible. This indicates that signal from the community structure is present in representations. It is especially true for \SINrNR that rely on community structure.

\subsubsection{Macroscopic-Level}
\label{subsec:exp_macroscopic_level}

\paragraph{PageRank score regression.} From a microscopic perspective, PageRank~\cite{brin_anatomy_1998} can be used to study the local properties of a vertex and its neighbors. As a local measure, it may help to gain insight into the flow of information or the influence within the network. From the macroscopic perspective, PageRank provides a measure of the importance of vertices in the network and may help to identify influential or central vertices that take part in the structure and function of the network. 

Similarly to the experiments in Section~\ref{subsec:exp_microscopic_level} and \ref{subsec:exp_mesoscopic_level}, we employ a linear regression with the objective of predicting the PageRank score of each vertex of the network from its vector representation. The average R$^2$ coefficient of determination over 50 runs is presented in Table~\ref{tab:pagerank_regression}.

\begin{table}[h!]
\centering
\resizebox{\textwidth}{!}{%
\begin{tabular}{@{}l|cc||ccccc@{}}
\toprule
         & \SINrNR                            & \SINrMF & \HOPE  & \FDWLK & \FWKLT & \FLouvainNE & \FVERSE  \\ \midrule
\COR     & \textbf{0.980} & -0.610 & 0.697 & -0.191   & -0.026   & -0.092    & 0.185  \\
\CTS & \textbf{0.949} & -0.176 & 0.346 & -0.050   & 0.182    & -0.123    & -0.125 \\
\EML    & \textbf{0.991} & 0.013  & 0.959 & 0.347    & -0.960   & -0.271    & 0.856  \\
\ARX    & \textbf{0.955}                           & 0.518  & 0.736 & 0.295    & 0.693    & 0.039     & 0.520  \\
\FB & \textbf{0.961}                           & 0.618  & 0.739 & 0.213    & 0.720    & 0.017     & 0.652  \\ \bottomrule
\end{tabular}%
}
\caption{PageRank vertex regression results: average $\text{R}^2$ over $50$ runs. }
\label{tab:pagerank_regression}
\end{table}

PageRank's regression results give a clear advantage to \SINrNR that outperforms all other methods. \HOPE is the second-best method for predicting PageRank, followed by \FVERSE. Except for the \FFB (\FB) network, \FDWLK, \FWKLT, \FLouvainNE and \SINrMF perform poorly on PageRank prediction. \SINrNR's domination on PageRank prediction may be correlated to its ability to predict degree. As a matter of fact, PageRank score is influenced both by the degree of a vertex at the microscopic level but also by the importance of its neighbors in the network. Both of these properties are at the heart of \SINrNR where we distribute the degree over the community structure using the neighbors of the vertex in different communities. Thus, the stronger a value for a community may be, the more it may contribute to the PageRank score of a vertex. Subsequently, \SINrNR and \HOPE to a lesser extent, can embed the PageRank, a node feature that relies on the global topology of the graph.

\subsection{{Application to word co-occurrence networks for word embedding}}\label{subsub:wordemb}

So far, all our experiments were performed on data naturally inclined to a graph representation. However, one of our main goals with this paper is to design a method that also allows representing words meanings from large corpora, i.e., extracting word embeddings. Furthermore, working with text extends the field of possibilities regarding intrinsic model evaluation—internal structure of the vector space—and also interpretability that is more easily auditable with words' semantics.

\paragraph{Word co-occurrence networks.} Word co-occurrence graphs are extracted from large textual corpora. To construct them, a sliding context window is applied over the sentences, allowing to sample the contexts of our vocabulary.

Let a weighted network $G=(V,E,\Omega)$. In our context, $V$ is the set of vertices representing words in set $L$ the lexicon extracted from the corpora. $E$ is the set of edges such that two vertices $u$ and $v$ representing two words $w_u$, $w_v \in L$ are connected when they appear in vicinity of each other within the context window. The edge weight $\omega_{u,v} \in \Omega$ is the count of how many times $w_u$ and $w_v$ have been observed together. 

To limit the number of vertices in the co-occurrence graph, we apply several preprocessing steps. To avoid preprocessing some words, we keep a list of exceptions, but we filter out words with less than three characters as a proxy of stop words (e.g., a, or) that do not carry sense. We also filter words appearing less than 20 times and chunk named entities under a single type and lowercase the vocabulary (e.g., ``\textit{Emperor Julius Caesar}'' becomes ``\textit{emperor\_julius\_caesar}''). Before constructing a co-occurrence graph from the co-occurrence matrix $\mathcal{A}$, we apply an additional filtering step. Using the \emph{Pointwise Mutual Information} measure (PMI), we remove co-occurrences which appear less than they would by chance. To do so, we calculate the PMI using the following rule for words $w_u, w_v \in L$ :

    \begin{equation}
        p(w_u, w_v) = \frac{cooc(w_u,w_v)}{\sum\limits_{w_i \in L}\sum\limits_{w_j \in L} cooc(w_i, w_j)}
    \end{equation}
        \begin{equation}
            p(w_u) = \frac{occ(w_u)}{\sum\limits_{w_i \in L} occ(w_i)}
        \end{equation}
    \begin{equation}
            \mathcal{A}_{(u,v)}=
       \begin{cases}
            0,~\text{if}~log\Big(\frac{p(w_u,w_v)}{p(w_u)\times p(w_v)}\Big) < 0 \\
            cooc(w_u, w_v),~\text{otherwise}
           
       \end{cases}
    \end{equation}
with $cooc(w_u,w_v)$ the number of co-occurrences between $w_u$ and $w_v$, $occ(w_u)$ the number of occurrences of $w_u$. In essence, the PMI acts as a threshold to limit the number of edges to only the relevant co-occurrences and helps to better detect communities. Apart from these preprocessing steps, the \SINrNR pipeline remains unchanged. We first detect communities and then measure the relative distribution of the degree of each vertex $u \in V$ representing a word $w_u \in L$ over the communities. By extension, we can see communities in co-occurrence networks as clusters of related words that co-occur frequently together and are thus semantically intertwined. This semantic relatedness among words within communities is at the heart of the interpretability of \SINrNR as we demonstrate in Section~\ref{subsec:exp_interpretability}.

\subsubsection{{Assessment of the performance of graph-based word embeddings}}

\paragraph{}Since text can also be represented with word co-occurrence graphs, we show the versatility of \SINrNR with the task of word embedding. As \SINrMF optimizations are hardly scalable for text (see Table~\ref{tab:runtime}), we will only consider \SINrNR for word co-occurrence networks. The word embeddings produced are then evaluated on classical intrinsic evaluation tasks for distributional models in \nlp : \textit{word similarity} and \textit{concept categorization}. We first introduce the process to create and embed words with \SINrNR, which is different from the one we described in \emph{Prouteau et al.}~\cite{prouteau_sinr_2021}: it runs faster, leads to robust representations (see Section~\ref{subsec:exp_robustness}), and using the $\gamma$  parameter, provides flexibility. Then, we evaluate the performances of this word embedding model against classical word embedding methods, some relevant graphs approaches and also a framework producing interpretable embeddings.

\paragraph{Experimental setup: data and competing approaches.}Co-occurrence networks can be constructed from any collection of texts. We employed two corpora that are composite in genre, open and readily available to the public.
\begin{itemize}
    \item[a.] \emph{Open American National Corpus} (\OANC) \cite{oanc_corpus} is the text part of the collection in contemporary American English, with texts from the 1990s onward. The corpus contains 11 million tokens prior to preprocessing and 20,814 types (vocabulary) for 4 million tokens after preprocessing.
    \item[b.] \emph{British National Corpus} (\BNC) \cite{bnc_oxford} is the written part of the corpora from a wide variety of sources to represent a wide cross-section of British English from the late 20$^\text{th}$ century. The raw corpus contains 100 million words. After preprocessing, the corpus is reduced to 40 million occurrences for a vocabulary of 58,687 types.
\end{itemize}

The baseline models we selected aim at representing a diversity of approaches to embedding words. Since our approaches leverage graph techniques to embed text, we also include two methods not specifically designed for word embedding but for graph embedding applied to word co-occurrence networks. The classical word embedding methods include \FWtoV \cite{mikolov_efficient_2013}, a pioneering model to embed words, \FSPINE, a method based on autoencoders to provide interpretable word embeddings and \SINrNR, our unified method able to embed both networks and words. The two graph embedding approaches applied to text we selected are: \HOPE, that is factorization-based—it is thus similar in philosophy to \Fglo~\cite{pennington_glove_2014}— and \FLouvainNE that derives embeddings from a hierarchy of communities obtained with \louvain—because it uses \louvain hierarchy. Approaches based on a combination of random walks and \FWtoV (\FWKLT and \FDWLK) are not relevant to this task: \FWtoV can be directly computed on the corpus.
 \FWtoV, \HOPE and \FLouvainNE are of dimension $300$ for both corpora. With \SINrNR, our 
interpretable models have respectively 4,460 and 8,454 dimensions on \OANC and \BNC~\cite{prouteau:hal-03770444}. The competing rival approach \FSPINE, has $1,000$ dimensions as advised by Subramanian et al.~\cite{subramanian_spine_2018}.

To evaluate performance of words' representations, we have on our hand classical intrinsic evaluation tasks. They are intrinsic in the sense that they rely solely on the embeddings in isolation over extrinsic which aims at evaluating the performances of a representation over a downstream task (e.g., \textit{text classification}, \textit{named entity recognition}, \textit{machine translation}). The first one is \textit{pairwise word similarity}.

\paragraph{Word similarity.} The \textit{word similarity} evaluation stems from the early work of \textit{Osgood et al.}~\cite{osgood_measurement_1957} and later \textit{Rubenstein and Goodenough}~\cite{rubenstein_contextual_1965} in psycholinguistics to test the distributional hypothesis. These experiments were later reintroduced by \textit{Baroni et al.}~\cite{baroni_dont_2014} as a way to evaluate the quality of distributional representations. The task relies on human constructed and annotated datasets containing pairs of words. Each pair of words is given a score by annotators: the higher the score, the more similar the two words presented are. For example, we could expect animals such as "elephant" and "cat" to be closer to each other than to the tool "hammer". The word similarity evaluation process is illustrated in Figure~\ref{fig:wordsim-example}. A score for a pair of words in the evaluation dataset is compared to the value of \textit{cosine similarity} for the same pair in the word embedding model. The \textit{Spearman Correlation} between the series of similarities given by humans and according to the cosine similarity in the model is then used as a quality metric for the vector space. Thus, the \textit{word similarity}  assess the quality of the neighborhood of the vectors: \textit{tiger} and \textit{lion}, or other words with a high similarity score, should indeed be close in the representation space. But it also evaluates the existence of a larger structure in the vector space. Since datasets are comprised of word pairs within a spectrum of scores, they allow observing some kind of distance between immediate neighborhoods : \textit{lion} and \textit{tiger} may be very close, but \textit{cat} should not be very far, and \textit{dog} may follow, or at least be much closer than \textit{hammer}.

\begin{figure}[h!]
\centering
\begin{tabular}{|c|c|c|c|c|}
\cline{1-3} \cline{5-5}
$w_1$     & $w_2$        & human rating &  & cosine\_sim($w_1$, $w_2$) \\ \cline{1-3} \cline{5-5} 
tiger & cat & 7.35 & \multirow{4}{*}{\begin{tabular}[c]{@{}c@{}}Spearman\\ Correlation\\ $\times$\end{tabular}} & 0.73 \\ \cline{1-3} \cline{5-5} 
plane  & car       & 6.31      &  & 0.65            \\ \cline{1-3} \cline{5-5} 
drink  & mother    & 2.85      &  & 0.20            \\ \cline{1-3} \cline{5-5} 
forest & graveyard & 1.85      &  & 0.12            \\ \cline{1-3} \cline{5-5} 
\end{tabular}%
\caption{Example of word similarity rating from the \men dataset and cosine similarity between vectors.}\label{fig:wordsim-example}
\end{figure}

Datasets used to evaluate word similarity are the following:
\begin{itemize}
    \item[a.] \men \cite{bruni_multimodal_2014} is composed of 3,000 pairs selected among the 700 most frequent words in \UKWAC and Wackypedia corpora. Ratings were collected on Amazon Mechanical Turk.
    \item[b.] \ws \cite{finkelstein_placing_2001} \emph{WordSim-353} contains 353 noun pairs.
    \item[c.] \scws \cite{huang_improving_2012} \emph{Stanford Contextual Word Similarity} dataset comprises 2003 pairs of mixed part-of-speech with senses sampled from WordNet \cite{miller_wordnet_1995}.
\end{itemize}
All three datasets comprise pairs of words both representing word \emph{similarity} (approximately synonymy, or at least substitutability like "cat" and "feline") and word \emph{relatedness} (much broader, encompasses pairs like "cup" and "coffee"). However, the datasets differ regarding the parts of speech included: \ws only includes nouns, while \men and \scws include nouns, verbs and adjectives.

\begin{table}[h!]
\centering
\begin{tabular}{@{}lccc@{}}
\toprule
\OANC  & \men                     & \ws                     & \scws                     \\ \midrule
\FWtoV  & \textbf{0.43                    }& \textbf{0.50}                   &\textbf{0.46}                     \\
\FSPINE & 0.33                    & 0.38                   & 0.44                     \\
\SINrNR  & 0.39                    & 0.44                   & 0.39                     \\
 \HOPE& 0.33& 0.43&0.39\\
 \FLouvainNE& 0.29& 0.37&0.23\\ \bottomrule
\BNC   & \men & \ws & \scws \\ \midrule
\FWtoV  & \textbf{0.72}                    & \textbf{0.65}                   & \textbf{0.57}                     \\
\FSPINE & 0.66                    & 0.57                   & 0.54                     \\
\SINrNR  & 0.66                    & 0.62                   & 0.54                     \\
 \HOPE& 0.53& 0.54&0.53\\
 \FLouvainNE& 0.28& 0.36&0.25\\ \bottomrule
\end{tabular}
\caption{Average Spearman correlation over 10 runs for \men, \ws and \scws word similarity datasets. }
\label{tab:similarity}
\end{table}

\begin{figure}[h]
    \centering
    \begin{subfigure}[t]{0.49\textwidth}
        \centering
        \includegraphics[width=\textwidth]{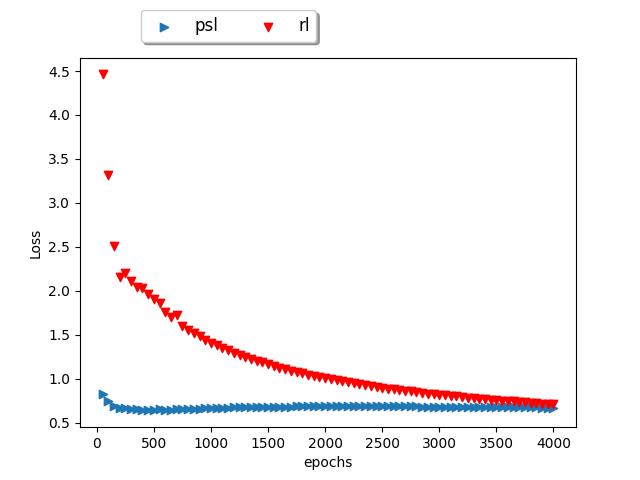}
        \caption{Losses on \OANC}
        \label{subfig:spineoanc}
    \end{subfigure}%
    ~ 
    \begin{subfigure}[t]{0.49\textwidth}
        \centering
        \includegraphics[width=\textwidth]{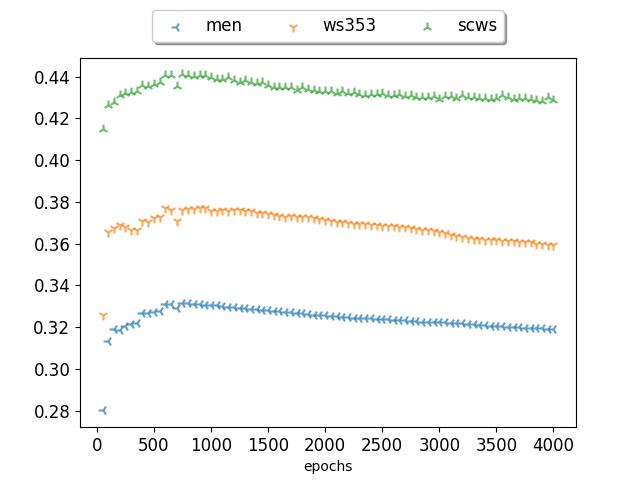}
        \caption{Similarity results on \OANC}
        \label{subfig:spineoancsim}
    \end{subfigure}
    ~
    \begin{subfigure}[b]{0.49\textwidth}
        \centering
        \includegraphics[width=\textwidth]{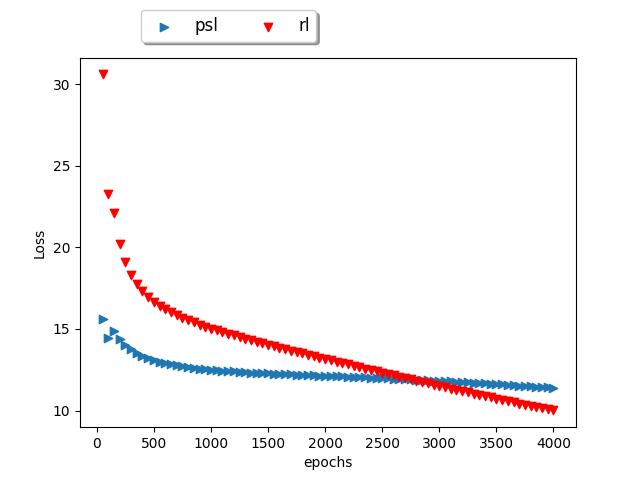}
        \caption{Losses on \BNC}
        \label{subfig:spinebnc}
    \end{subfigure}
    \begin{subfigure}[b]{0.49\textwidth}
        \includegraphics[width=\textwidth]{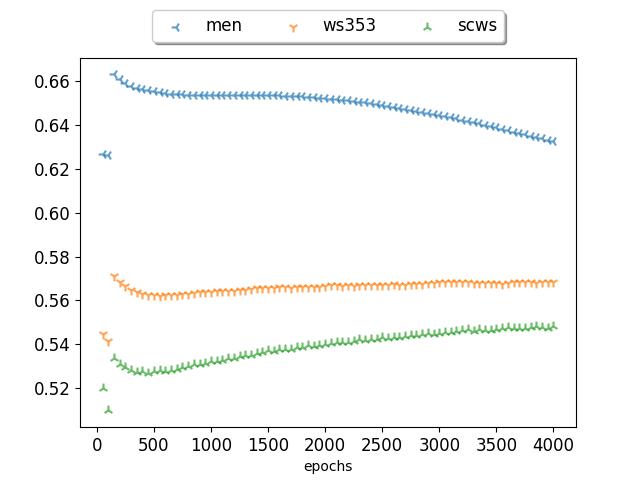}
        \caption{Similarity results on \BNC}
        \label{subfig:spinebncsim}
    \end{subfigure}
    \caption{Average losses and similarity results for \FSPINE for 10 runs on 4k epochs. psl stands for Partial sparsity loss that controls interpretability, rl for reconstruction loss that controls performance.}
    \label{fig:spine-sim}
\end{figure}

\textit{Word similarity} values in Spearman correlation are averaged over 10 models trained with each corpus and presented in Table~\ref{tab:similarity}. For \FSPINE, the best results regarding the $4k$ epochs of training are kept, as shown in Figure~\ref{fig:spine-sim}. The average correlation values are rather close across models and datasets. The gap between interpretable models—\FSPINE and \SINrNR—and \FWtoV is at most of 7 points. The gap is even tighter on our larger corpora \BNC. \SINrNR and \FSPINE remain very close to \FWtoV. These results strongly suggest that one can build interpretable representations that still perform close to dense embedding models such as \FWtoV. Regarding graph-based approaches, \HOPE also performs very well on \OANC with results similar to those of \FSPINE. However, it does not scale up with the higher number of co-occurrences of \BNC, results are lower than those of \FSPINE and \SINrNR. \FLouvainNE achieved good performances on other tasks, but on this task, results may not reflect the true performances of the algorithm, its implementation being only able to deal with unweighted graphs. The textual graphs, being weighted, \FLouvainNE is disadvantaged.

Results achieved by \SINrNR on a text-oriented task thus show its versatility regardless of the type of information modeled by the network. \SINrNR results are indeed better than those of \FSPINE on \men and \ws, the other interpretable approach, and are always better than those of \HOPE and \FLouvainNE, the other graph-based approaches.

Furthermore, if we consider the run-time as shown in Table~\ref{tab:runtimewords}, \SINrNR results are also convincing. Even if the approach is slower than \FWtoV on \BNC, it actually requires less compute (\textsc{CPU} time). And we shall recall that the total run-time of \SINrNR is the sum of times required to extract the co-occurrence network and to train the embeddings. It is thus interesting to note that for $10$ runs of \SINrNR (for instance to tune $\gamma$), the co-occurrence matrix run-time is considered only once, and it is thus actually faster than \FWtoV (3,658s for \SINrNR and 4,890s for \FWtoV). When compared to \FSPINE (4k epochs as preconised by the authors), the competing interpretable approach, \SINrNR is far more time-efficient. First, \FSPINE uses dense vectors such as \FWtoV and make them interpretable, its run-time thus includes the one of \FWtoV. Furthermore, the k-sparse auto-encoding of the dense vectors to make them interpretable requires quite some time. For \OANC, the total CPU time required to run the 4k epochs of \FSPINE is $1,000$ times the one required for \SINrNR. In Figure~\ref{fig:spine-sim}, the losses seem to indicate that 4,000 epochs are required, but similarity results may indicate that 1,000 is enough. In such a case, \FSPINE's run-time would be divided by four, which still makes it above the runtime of \SINrNR, our interpretable approach, by a large margin. If one considers run-times of the competing graph approaches on the textual data in Table~\ref{tab:runtimegraphs}, the results are also interesting. \FLouvainNE is the fastest approach, but it does not succeed in achieving good results for this similarity task. \HOPE which is the best competing graph approach regarding similarity results, runs ten times slower than \SINrNR, demonstrating the ability of \SINrNR to deal with graphs and textual data efficiently.

\begin{table}[!ht]
    \centering
    \begin{tabular}{|p{3cm}|c|c|c|c|}
    \hline
        & \multicolumn{2}{|c|}{\OANC} & \multicolumn{2}{|c|}{\BNC} \\ \cline{2-5} 
        ~ & \textsc{cpu} time & run-time & \textsc{cpu} time & run-time \\ \hline
        \FWtoV & 175 & 49 & 1805 & \textbf{489} \\ \hline
        \FSPINE & 50k & 15k & 130k & 36k \\ \hline
        Graph extraction & 6 & 7 & 188 & 188 \\ 
        + \SINrNR training & 38 & 34 & 383 & 347 \\ 
        = \SINrNR total & \textbf{44} & \textbf{41} & \textbf{571} & 535 \\ \hline
    
    \end{tabular}
    \caption{Average total CPU time and run-time in seconds over 10 runs. Run-time is computed with four cores on \emph{Intel Xeon E5-2690 v2 3.00GHz CPU}. For \FSPINE, "k" indicates kilo-seconds. \SINrNR total is the sum of \SINrNR training (\SINrNR column) and of the co-occurrence matrix computing (co-occ column). }
    \label{tab:runtimewords}
\end{table}

\begin{table}[!ht]
    \centering
    \begin{tabular}{|p{3cm}|c|c|c|c|}
    \hline
        & \multicolumn{2}{|c|}{\OANC} & \multicolumn{2}{|c|}{\BNC} \\ \cline{2-5} 
        ~ & \textsc{cpu} time & run-time & \textsc{cpu} time & run-time \\ \hline
        \SINrNR & 38 & 34 & 383 & 347 \\ \hline
        \FLouvainNE  & \textbf{3} & \textbf{3} & \textbf{16} & \textbf{16} \\ \hline
        \HOPE  & 492 & 425 & 3799 & 3748 \\ \hline
    
    \end{tabular}
    \caption{Average total CPU time and run-time in seconds over 10 runs. Run-time is computed with four cores on \emph{Intel Xeon E5-2690 v2 3.00GHz CPU}. The time required to compute the co-occurrence graph is not included, figures only reprent the training time.}
    \label{tab:runtimegraphs}
\end{table}

To further evaluate \SINrNR's abilities to model text data, we now consider a \textit{concept categorization} evaluation.

\paragraph{Concept categorization.} Concept categorization or word clustering is the text pendant of our community clustering approach previously presented. Instead of communities, the goal of concept categorization is to properly cluster a subset of selected words from their embedding into preset categories. Since the categories encompass basic-level concepts (\textit{cat} and \textit{dog} opposed to \textit{golden retriever} and \textit{german shepherd}), their assessment implies the existence of a larger structure than immediate proximity for substitutable words : topical consistency in regions of the representation space.  Categories in datasets range from animals to feelings or legal documents to cite only a few. The datasets we chose, include:

\begin{itemize}
    \item[a.] \AP \cite{almuhareb_concept_2005} is a categorization dataset constructed with the goal of being balanced in class type, term frequency and ambiguity. The dataset contains 21 different categories.
    \item[b.] \BLESS \cite{baroni_how_2011} contains 200 nouns (100 animate, 100 inanimate) from 17 categories (e.g. appliance, bird, vehicle, vegetable).
    \item[c.] \ESSLLI \cite{esslli_dataset_2008} datasets have been created for the shared tasks of \emph{Workshop on Distributional Lexical Semantics} that took place during the 2008 \FESSLLI. Three datasets for concepts categorization were constructed with regard to three tasks. \ESSLLIa aims at grouping $44$ nouns into semantic categories (4 animate, 2 inanimate). \ESSLLIb focuses on categorizing $40$ nouns in three concreteness levels: low, moderate, high. \ESSLLIc evaluates the clustering of $45$ verbs into $9$ categories.
\end{itemize}

Clustering is operated on the embedding vectors of the words in each dataset provided by each method. Words are thus clustered into categories from their vectors using the \emph{K-means} and \emph{hierarchical} algorithms, only the best purity is retained and averaged. Concept categorization results are reported in Table \ref{tab:cat-oanc} and \ref{tab:cat-bnc}. Clustering performance is measured in purity between ground truth clusters in datasets and detected clusters.

\begin{table}[h!]
\centering
\begin{tabular}{@{}l|c||cc||cc@{}}
\toprule
\OANC      & \SINrNR  & \FSPINE & \FWtoV & \HOPE  & \FLouvainNE \\ \midrule
\AP        & 0.299 & 0.325 & \textbf{0.353}    & 0.233 & 0.186     \\
\BLESS     & 0.402 & 0.376 & \textbf{0.411}    & 0.276 & 0.225     \\
\ESSLLIc & \textbf{0.489} & 0.444 & 0.469    & 0.431 & 0.378     \\
\ESSLLIb & 0.688 & 0.680 & \textbf{0.702}    & 0.615 & 0.635     \\
\ESSLLIa & 0.516 & 0.573 & \textbf{0.593}    & 0.457 & 0.475     \\ \bottomrule
\end{tabular}
\caption{Concept categorization purity scores over 10 runs for \OANC.}
\label{tab:cat-oanc}
\end{table}

\begin{table}[h!]
\centering
\begin{tabular}{@{}l|c||cc||cc@{}}
\toprule
\BNC                            & \SINrNR  & \FSPINE & \FWtoV & \HOPE  & \FLouvainNE \\ \midrule
\AP                             & 0.541 & 0.567 & \textbf{0.589}    & 0.396 & 0.183     \\
\BLESS                          & 0.755 & 0.774 & \textbf{0.832}    & 0.455 & 0.357     \\
\ESSLLIc                      & 0.580 & 0.538 & \textbf{0.594}    & 0.489 & 0.329     \\
\ESSLLIb                      & 0.708 & 0.703 & 0.700    & \textbf{0.738} & 0.675     \\
\ESSLLIa & 0.786 & 0.732 & \textbf{0.798}    & 0.630 & 0.543     \\ \bottomrule
\end{tabular}
\caption{Concept categorization purity scores over 10 runs for \BNC.}
\label{tab:cat-bnc}
\end{table}

Categorization results on \OANC, our smallest dataset show close purity results between all methods, although \FWtoV is leading on most datasets by a short margin. Methods not necessarily dedicated to the task of word embedding perform lower. Overall, the purity scores on our smaller corpora are lower than on \BNC which contains more occurrences. On \BNC, the trend is the same as on \OANC: \SINrNR, \FSPINE and \FWtoV are very close to one another. \HOPE and \FLouvainNE have subpar performances except for \HOPE on \ESSLLIb for word concreteness level categorization.

\subsubsection{{Assessment of the stability of representations}}
\label{subsec:exp_robustness}

\paragraph{}Interpretability is one of the main objectives of \SINr.
With interpretability comes the ability to audit a model and make conjectures based on its internal organization. Hypothesizing based on the internal structure of a vector space is easier when the method used to embed data is relatively stable across runs. Stability can either be in terms of neighbors in the embedding space, meaning that the geometry of the projection space remains unchanged, or in the case of \SINrNR directly related to invariance in the community structure. To investigate the stability of embedding models, we study Section~\ref{subsec:exp_robustness} the robustness of multiple methods. 

\emph{Pierrejean}~\cite{pierrejean:tel-02628954} investigated the variance of neural methods to derive word embeddings. \FWtoV is notoriously unstable across runs and word neighborhoods may be distorted with consequences on \emph{Word Similarity Evaluations}. Instability impedes interpretability in the sense that we are not guaranteed to get the same model across runs. To measure variability in models, we first evaluate the stability of \SINrNR's \louvain community detection, since it is the only random process in our algorithm. In a second evaluation, we take a look at the variation of neighbors between models.

\paragraph{Community structure stability.} Inconsistency in \SINrNR vectors may stem from \louvain's random iteration on vertices. As a result, communities may change between instances of \SINrNR. Subsequently, with a change in community structure comes a change in the representation of items. To that end, we evaluate the variation in community structure for 10 community structures that allowed to extract \SINrNR vectors. We evaluate the pairwise \emph{Normalized Mututal Information} (NMI) and present the averaged NMI for all pairs of community structures in Table~\ref{tab:nmi-communities}.

\begin{table}[h!]
\centering
\begin{tabular}{@{}lcc@{}}
\toprule
    & \OANC  & \BNC   \\ \midrule
NMI & 0.967 & 0.959 \\ \bottomrule
\end{tabular}
\caption{Average NMI comparing 10 community structures detected with \louvain on \OANC and \BNC co-occurrence networks.}
\label{tab:nmi-communities}
\end{table}

NMI values are high, meaning that despite randomness in \louvain order of iteration over the vertices, community detection leads to similar partitions of the vertices. Preprocess described in Section~\ref{subsub:wordemb} using PMI filtering may explain this. Similar partitions should lead to little variation in embedding space geometry. More precisely, words neighborhoods in \SINrNR should not vary too much for two models extracted from the same network. This is the subject of our next experiment on word nearest neighbors variation. 

\paragraph{Word neighborhood variation.} We have previously seen that \SINrNR communities vary by a small extent between instances of two models. Variation in neighborhoods was demonstrated for other word embedding methods \cite{pierrejean:tel-02628954}. To evaluate this variation, we do a pairwise comparison of word neighbors for 10 models. The \emph{Nearest Neighbor Variation} \cite{pierrejean:tel-02628954} described in Equation~\ref{eq:varnn} measures the proportion of varying nearest neighbors between two models $M_1$ and $M_2$ for a number $N$ of nearest neighbors $nn$ according to the cosine distance. For a word $w$, the \emph{Nearest Neighbor Variation} ($varnn$) score is:
\begin{equation}
    varnn_{M_1, M_2}^{N}(w) = 1 - \frac{|nn_{M_1}^{N} \cap nn_{M_2}^{N}|}{N}
    \label{eq:varnn}
\end{equation}

\begin{figure}[h!]
    \centering
    \begin{subfigure}[t]{0.5\textwidth}
        \includegraphics[width=\textwidth]{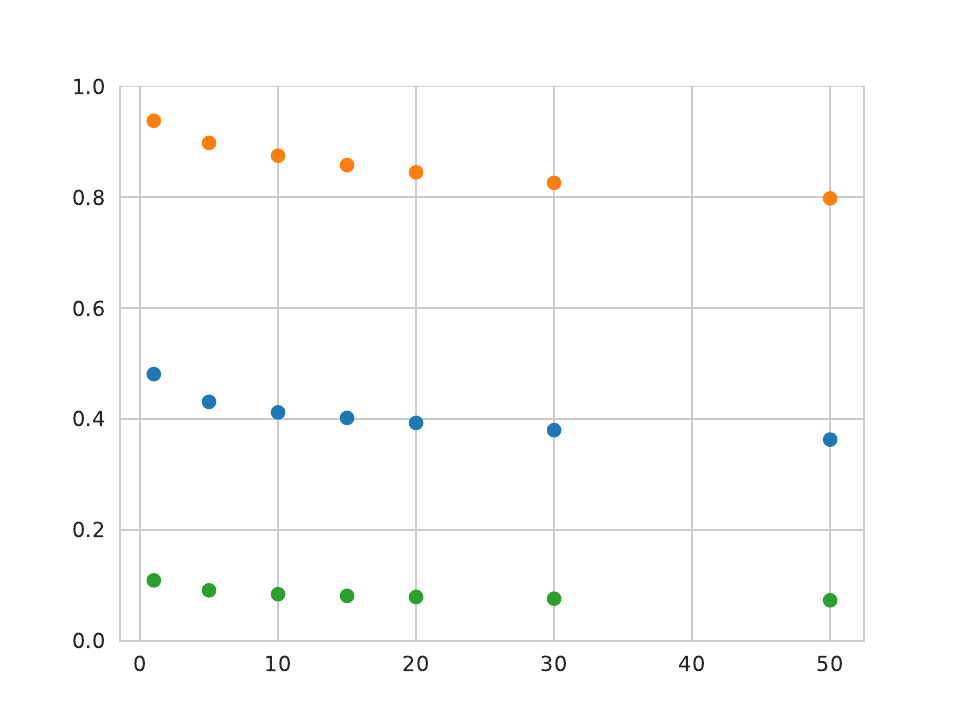}
    \end{subfigure}%
    \hfill
    \begin{subfigure}[t]{0.5\textwidth}
        \includegraphics[width=\textwidth]{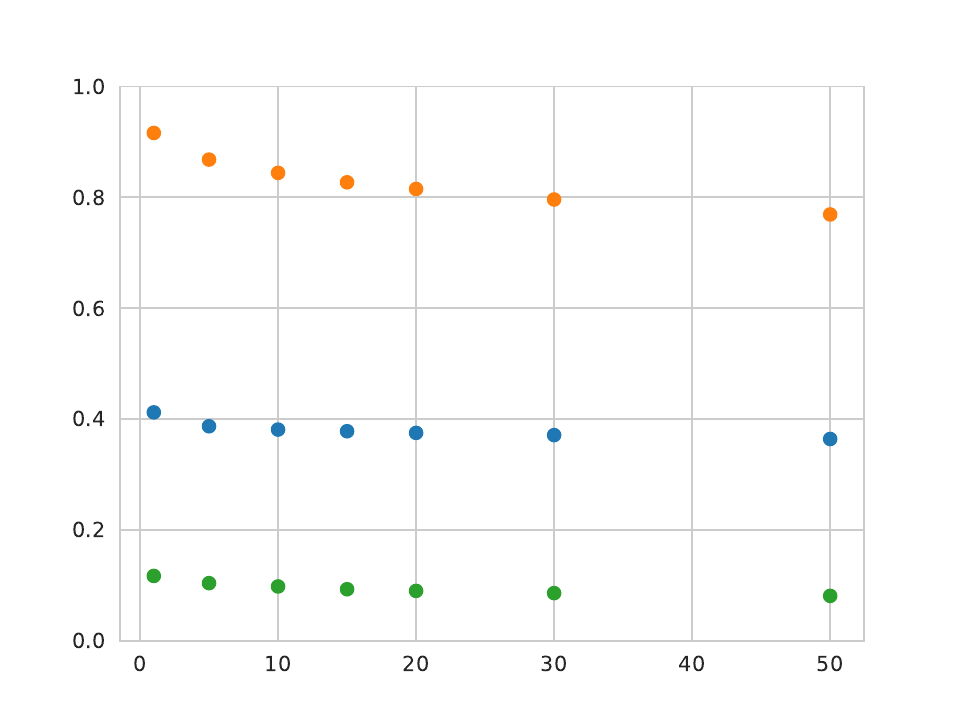}
    \end{subfigure}
    \caption{Neighborhood instability (average $varnn$) according to the number of nearest neighbors in cosine similarity for \OANC (left) and \BNC (right). Instability of: \FSPINE (orange) topmost values, \FWtoV (blue) middle values and \SINrNR (green) bottom most values for a pairwise comparison of 10 models.}
    \label{fig:stability-oanc-bnc}
\end{figure}

In Figure~\ref{fig:stability-oanc-bnc}, we can see the variation in instability $varnn$ according to the distance at which we are retrieving nearest neighbors, i.e., $N$ in Equation~\ref{eq:varnn}. First and foremost, \SINrNR's neighbors variation between models is weak both on \OANC and \BNC. \FWtoV is right in between \SINrNR and \FSPINE in terms of variation with values between $0.38$ and $0.50$. \FSPINE's word embedding neighbors vary a lot more, with variation proportions being mostly over $0.8$. The instability in \FSPINE could be expected, \FSPINE is derived from a dense \FWtoV space, which varies. Nearest neighbors remain similar for \SINrNR even at a distance of 50. On the other hand, \FSPINE and \FWtoV's nearest neighbor strongest variation seems to be located within the first 20 nearest neighbors. 

Variation in neighbors between models can be detrimental to interpretability, as hypotheses drawn from one model are not necessarily true on another model trained with the same algorithm and the same data. Subsequently, stable representations are preferable for applications requiring to be audited, in the context of digital humanities~\cite{gefen2017vector}, and when using diachronic alignments~\cite{hamilton2016cultural,garg2018word}.

\subsection{Interpretability}
\label{subsec:exp_interpretability}

After having thoroughly evaluated the performance of our approach for vertex embedding and word embedding, we study the interpretability associated with the models produced. This is one of the main perks of our approach with its low compute, we carefully evaluate it on text, since it is easy for humans to interpret text labels.

\paragraph{Word intrusion detection.}The question of interpretability is intertwined with human perception of dimensions' coherence. Subsequently, an evaluation task specifically designed to evaluate dimension interpretability first appeared in \textit{Chang et al.}~\cite{chang_reading_2009} to evaluate the coherence of words describing topics in \textit{topic models}. The \textit{word intrusion} evaluation task aims at assessing the extent of a models' dimension interpretability and has become the \textit{de facto} evaluation of interpretability \cite{murphy_learning_2012, faruqui_sparse_2015, subramanian_spine_2018, prouteau:hal-03770444}. 

The task is based on a simple principle: if a vector space is well structured, words that are semantically close should lie close together. This is the distributional hypothesis. Now, for words to be close in a space, chances are that their representation relies on common dimensions. That is where the \textit{word intrusion} becomes useful. If we select a dimension of the vector space and rank the words according to their value on this dimension, according to our precedent hypothesis, words with the strongest values should be semantically related. Now, how can be assured that the words are related? 

We use an \textit{"intruder"}, a word selected at random among those with the lowest values on our dimension of interest, but that is still strong on another dimension—to avoid picking too specific or rare of a word. If a native speaker of a language can find the intruder among this set of words, then the top scoring word of the dimension must possess some semantic consistency. This semantic consistency for dimensions corresponds to interpretability for word embeddings.

We evaluate two models with such \textit{Word Intrusion} protocol. The experiment is, as far as we know, the first of its kind on a large French corpus. Our models (\FSPINE, \SINrNR) were trained on a news corpus in French, it contains articles from the news outlet \textit{Le Monde} (1987-2006), \textit{AFP} (1994-2006) and news articles crawled on the web (2007). The text is purposely lemmatized, named entities are chunked under a single type and stop words are removed along with words occurring less than $10$ times. Most named entities were removed to construct intrusion tasks as the corpora spans multiple decades, which would rely too much on annotators' general knowledge instead of semantics. The preprocessed corpus contains $330$M tokens and $323$K words. We train a \FSPINE model which has 1,000 dimensions and \SINrNR with 4,708 dimensions. 

\begin{figure}[h!]
    \centering
    \includegraphics[width=.5\textwidth]{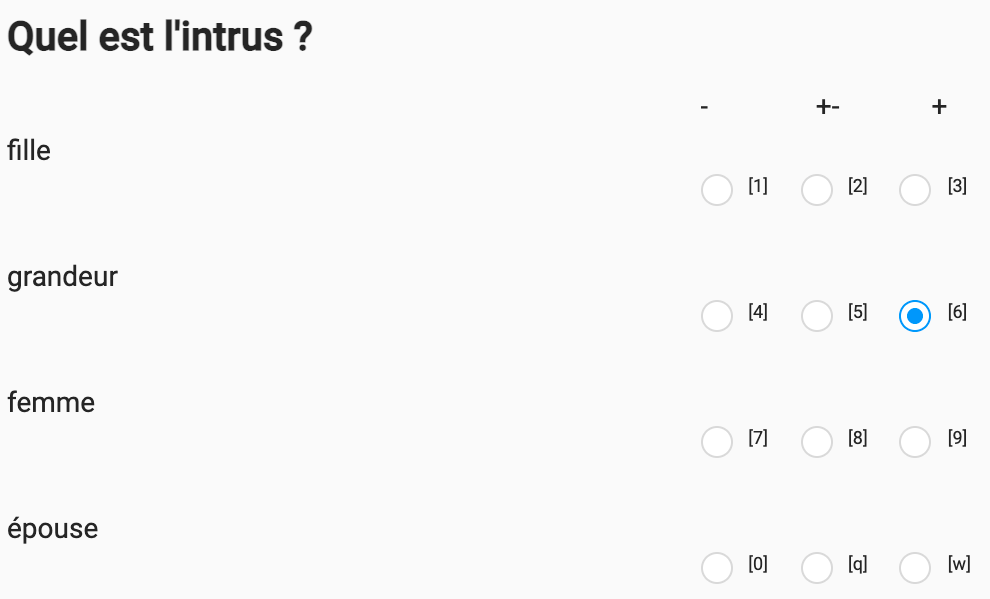}
    \caption{Example of a word intrusion task annotated. Among words: \textit{"daughter"}, \textit{"size"}, \textit{"woman"}, \textit{"wife"}. The intruder is \textit{"height"}.}
    \label{fig:intrusion-example}
\end{figure}

In our \textit{Word Intrusion} protocol, each task is extracted as follows: first, we sample a dimension. 
Then, we select the top 3 words having the highest values on the coordinate corresponding to this sampled dimension. We also sample an intruder which is part of the lower 30\% of values in the coordinate and in the top 10\% of another coordinate.  In total, 200 dimensions were sampled for \FSPINE and \SINrNR. For each word in the intrusion test, three possibilities are presented: ($-$, $\pm$, $+$). When annotators can easily detect the intruder, they should select $+$. When annotators hesitate between two words, they should select $\pm$. If the annotators find all the words consistent—everything seems coherent—they should select $-$. This allows to analyze finely the interpretability of dimensions—hesitations, all coherent words or no coherence. The intrusion tasks were served through a web interface on a \textit{Label-Studio} server. Table~\ref{tab:task_examples} presents a selection of intrusion tasks, and in Figure~\ref{fig:intrusion-example} we present an example of an intrusion task in the interface.

\begin{table}[h!]
\centering
\begin{tabular}{@{}ccccc@{}}
\toprule
Model &
  \multicolumn{3}{c}{Top Words} &
  Intruder \\ \midrule
\multirow{2}{*}{\FSPINE} &
  suffrage &
  \begin{tabular}[c]{@{}c@{}}urne\\ \textit{(ballot box)}\end{tabular} &
  \begin{tabular}[c]{@{}c@{}}législative\\ \textit{(legislative)}\end{tabular} &
  \begin{tabular}[c]{@{}c@{}}colmatage\\ \textit{(sealing)}\end{tabular} \\ \cmidrule(l){2-5} 
 &
  tramway &
  \begin{tabular}[c]{@{}c@{}}ferroviaire\\ \textit{(rail)}\end{tabular} &
  rail &
  orientation \\ \midrule
\multirow{2}{*}{\SINrNR} &
  \begin{tabular}[c]{@{}c@{}}réseau\\ \textit{(network)}\end{tabular} &
  \begin{tabular}[c]{@{}c@{}}chaîne\\ \textit{(channel)}\end{tabular} &
  \begin{tabular}[c]{@{}c@{}}groupe\\ \textit{(group)}\end{tabular} &
  \begin{tabular}[c]{@{}c@{}}déclencher\\ \textit{(trigger)}\end{tabular} \\ \cmidrule(l){2-5} 
 &
  Intel &
  \begin{tabular}[c]{@{}c@{}}microprocesseur\\ \textit{(microprocessor)}\end{tabular} &
  \begin{tabular}[c]{@{}c@{}}processeur\\ \textit{(processor)}\end{tabular} &
  \begin{tabular}[c]{@{}c@{}}garder\\ \textit{(to keep)}\end{tabular} \\ \bottomrule
\end{tabular}%

\caption{Examples of tasks extracted for each model.}
\label{tab:task_examples}
\end{table}

The pool of annotators is composed of $19$ master students in \nlp. The participants have prior knowledge of distributional models and are literate in French, with at least 4 months in France. They evaluated 66 tasks in random order. Each task was solved by three or four participants.

\begin{table}[h!]
\centering
\begin{tabular}{lcc}
\toprule
                                     & \FSPINE & \SINrNR  \\ \midrule
\intruderok                          & \textbf{36\%}   & 35\%           \\ 
+ \hesistateok                 & 56\%    & \textbf{60\% }        \\ 
+ \consistent & 57\%    & \textbf{62\%}       \\ \bottomrule

\end{tabular}
\caption{Positive results of the intrusion detection task. }\label{tab:posresults}
\end{table}

 In Table~\ref{tab:posresults} we present the percentages of tasks for which the intruder was correctly detected \intruderok, the \hesistateok category corresponds to a hesitation between two words in which the intruder is one of them. The \consistent category corresponds to tasks for which annotators found consistency in terms of sense across all the words in the task. Percentages are cumulative. 
First off, what we can see is that \FSPINE and \SINrNR are shoulder to shoulder on the \intruderok front. The percentages remain low, which indicates that the task is hard. In their paper, \textit{Subramanian et al.}~\cite{subramanian_spine_2018} operated a similar experiment, but on a reduced vocabulary size of 15k words, which is much lower than our $323$k words. In such a case, \FWtoV obtains a score of $26\%$ that should be compared to the $62\%$ of \SINrNR. We can already see that by considering \intruderok, \FSPINE and \SINrNR are ahead of \FWtoV on a smaller English vocabulary. Furthermore, when we consider the \intruderok + \hesistateok + \consistent, we can see that \SINrNR performs better and has more \consistent cases. It might be the consequence of the larger number of dimensions that leads to redundancy in the dimensions. 

In Table~\ref{tab:negresults} we analyze further \textit{word intrusion} evaluation results. We can see that \SINrNR has fewer instances where the annotators were quite sure about an intruder but failed to predict the right one (\intruderko). On the other hand, \FSPINE manages to have fewer instances where subject hesitated between two words and none of them was the intruder (\hesistateko). Lastly, there seems to be fewer instances for \SINrNR where subject were not able to discern a semantic coherence among the words they were presented with (\skipped).

\begin{table}[h!]
\centering
\begin{tabular}{lccc}
\toprule
                                     & \FSPINE & \SINrNR \\ \midrule
\intruderko                 &  14\%  & \textbf{12\%}            \\
\hesistateko                & \textbf{10\%}   & 11\%            \\
\skipped                    & 19\%   & \textbf{15}\%            \\ \bottomrule

\end{tabular}
\caption{Negative results of the intrusion detection task.}
\label{tab:negresults}
\end{table}

\begin{table}[h!]
\centering
\begin{tabular}{ccc}
\toprule
\FSPINE              & \SINrNR \\ \midrule
\textbf{58\%, 21\%} &  55\%, 13\% \\ \bottomrule
\end{tabular}
\caption{Inter-annotator agreements across all models presented and overall for the \textit{word intrusion} evaluation. For each model, the first value is the percentage of tasks where at least two evaluators annotated similarly. The second value is the percentage of tasks where the three evaluators annotated similarly.}
\label{tab:agreement}
\end{table}

Regarding inter-annotator agreement, \FSPINE has the highest agreement, in 58\% of cases, at least two annotators agree on their decision. \SINrNR is just behind with 55\%. When we consider the three annotators, the percentages of agreement drop significantly, \FSPINE is leading with annotators agreeing 21\% of the time. \SINrNR has a lower three annotators agreement with only 13\%. We also computed \textit{Fleiss' kappa}: \FSPINE has a 0.26 $\kappa$ and \SINrNR a 0.21 $\kappa$. These agreements fall in the \textit{fair agreement} category. The low $\kappa$ scores further confirm the difficulty of the \textit{Word Intrusion} detection task. Furthermore, we voluntarily left more choices than the original evaluation task, thus potentially lowering the potential for agreement.

To conclude on the \textit{word intrusion} detection, we have shown that there is potential for interpretability of word embeddings in \FSPINE and \SINrNR despite the complexity to evaluate. A more systematic evaluation of models' dimensions seems unlikely until we develop more automatic ways of evaluating these models, as \emph{Lau et al.}~\cite{lau_machine_2014} introduce.

The potential of these models for visualization of dimensions and embeddings has been thus far underexploited in this paper. We demonstrate in the next paragraph how interpretability can be used to probe word representation.

\begin{table}[h!]
    \centering
    \resizebox{\textwidth}{!}{
    \begin{tabular}{lll}
        \toprule
               & \multicolumn{1}{c}{\FSPINE}                                                                                                                                                                       & \multicolumn{1}{c}{\SINrNR}                                                                                                                                                                                                        \\ \midrule
        insulin &     \begin{tabular}[c]{@{}l@{}}glutathione, pancreas, gastroduodenal\\ immunologically, hyperplasia, transgene\\ insulin, sulphasalazine, interferon\end{tabular} & \begin{tabular}[c]{@{}l@{}}hypertriglyceridaemia, mellitus, porcine\\ aldosterone, aminotransferase, creatinine\\ ulcerative, sulphasalazine, colitis\end{tabular} \\ \midrule
        mint    &  \begin{tabular}[c]{@{}l@{}}spoonfuls, parsnips, kebabs\\ onion, basil, yogurt\\ dial, screams, vibration\end{tabular}                    & \begin{tabular}[c]{@{}l@{}}tbsp, oregano, diced\\ Gibson, gigged, charvel\\ minted, minting, hoards\end{tabular}                                                         \\ \bottomrule
    \end{tabular}%
    }
    \caption{Three (one per line) most activated dimensions for words "\textit{insulin}" and "\textit{mint}" and for each dimension, three words with the strongest values on each dimension.}
    \label{tab:strongest-dims}
\end{table}

\paragraph{Probing word sense.} Interpretable embedding models lend themselves naturally to visual interpretability. As we demonstrated in Section~\ref{sec:visual-introduction}, we can visualize vectors of airports to better understand their position within the network. Similarly, on word co-occurrence networks, and with the help of interpretable models, we can visualize how our vector space is structured. We observed with the \emph{word intrusion detection} that well-formed vector spaces should exhibit dimensions on which words with the strongest values are semantically related. A word's representation is dependent on a subset of dimensions which contribute to different extents to its representation. For instance, a sofa might be represented by a dimension representing the fact that it is a seat, another one related to fabric, a third one related to the living room etc. We can simply visualize the strongest coordinates for words and the strongest words for these coordinates, as in the \emph{word intrusion detection} tasks. In Table~\ref{tab:strongest-dims}, we have selected the three strongest dimensions for words \textit{"insulin"} and \emph{"mint"} on a \FSPINE model and also on a \SINrNR model, both trained on \BNC. The first thing that we can see is that the words on the top dimensions make sense with one another. Naturally, \emph{"insulin"} is largely represented by medical and physiological terms. More interestingly, \emph{"mint"} has more variety in terms of topics, especially for \SINrNR. Indeed, the first dimension is related to herbs and cooking. The second dimension is more surprising, since it seems to be about guitar brands. However, we know that \BNC contains classified ads for the sale of guitars and \emph{"mint"} is an adjective related to the condition of an object that may be used in classified ads. The third-strongest dimension is related to the monetary aspect of \emph{"mint"} and terms related to minting money. This example shows the polysemy of the term that can be captured from the co-occurrences in the corpus.
  
\begin{figure}[h!]
    \centering
    \includegraphics[width=\textwidth]{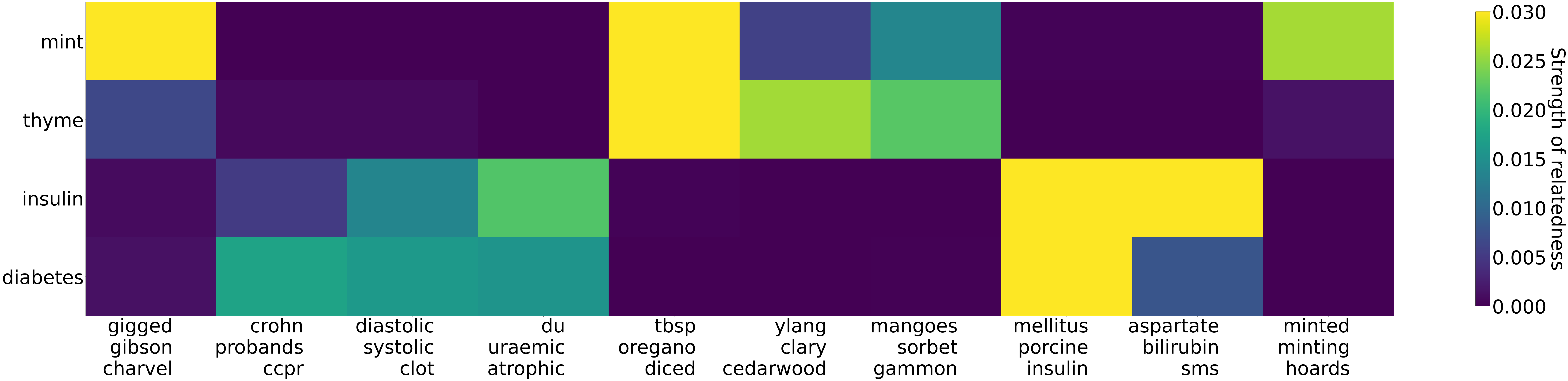}
    \caption{Common dimensions labeled with their strongest words (horizontal axis) for four words (vertical axis): \emph{"mint"}, \emph{"thyme"}, \emph{"insulin"} and \emph{"diabetes"} in a \SINrNR model trained on \BNC.}
    \label{fig:shared-dim-sinr}
\end{figure}

Interpretability can also be exploited by visualizing common dimensions for multiple words. For example, we would expect \emph{"cat"} and \emph{"dog"} to share dimensions but not \emph{"cat"} and \emph{"hammer"}. In Figure~\ref{fig:shared-dim-sinr} we present shared dimensions for words: \emph{"mint"}, \emph{"thyme"}, \emph{"insulin"} and \emph{"diabetes"}. The first thing we can see is that herbs do not seem to share strong dimensions with diabetes-related terms. Secondly, "thyme" and "mint" share the same dimensions as in Table~\ref{tab:strongest-dims} but also dimensions about herbs and wood sought after for their essential oils (\emph{"ylang"}, \emph{"clary"}, \emph{"cedarwood"}). Dimensions related to \emph{"diabetes"} and \emph{"insulin"} are a mix of cardiovascular terms (\emph{"diastolic"}, \emph{"systolic"}, \emph{"clot"}) and words related to the production of \emph{"insulin"} (\emph{"melitus", "porcine", "insulin"}). We see that similar words share dimensions. 

\begin{figure}[h!]
    \centering
    \includegraphics[width=\textwidth]{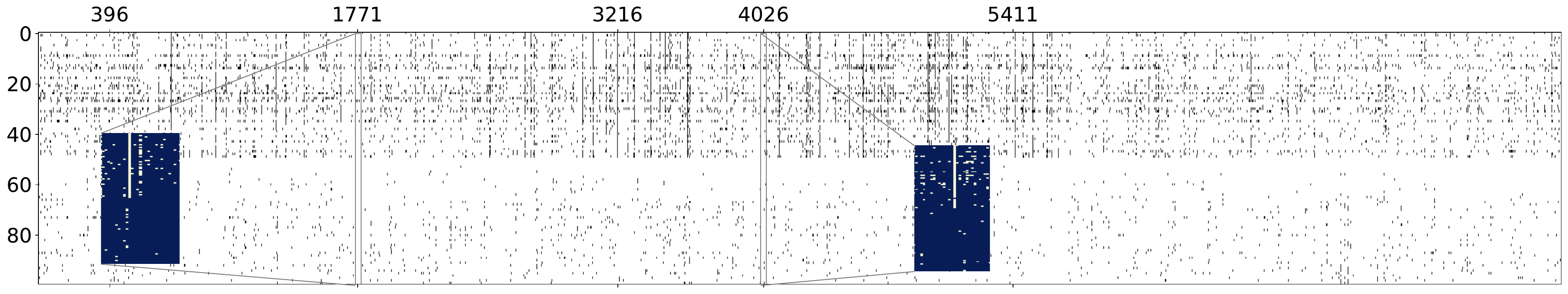}
    \caption{Visualizing shared dimensions, vectors of 50 closest (top) and most distant (bottom) words to \emph{"mint"} in a \SINrNR model trained on \BNC. The two insets are dimensions on which all 50 closest neighbors have values (white line). Strongest words on coordinate 1771: [\textit{"tbsp", "oregano", "diced", "dijon"}]; 4026: [\textit{"rind", "juice", "lemon", "cayenne"}]. }
    \label{fig:neighbors-shared-dimensions}
\end{figure}

We visualize shared dimensions at larger scale in Figure~\ref{fig:neighbors-shared-dimensions} where we plot the most (top half) and least (bottom half) similar words according to cosine similarity based on the vector of \emph{"mint"}. For visualization purposes, we only plot the presence or absence of a value for the dimension and not the magnitude of the value.
We can see lines appearing at the top of the figure, which indicates that dimensions are shared across neighbors in the vector space. We have zoomed in on two dimensions of the matrix for which most 50 closest neighbors have values. We can see a clear line on these inset, meaning that this shared dimension is used to characterize \emph{"mint"} and its close neighbors. Furthermore, among the word with the highest values on these components we find words related to cooking, citrus fruits and spices.

Interpretability is useful in applications where we might wish to understand how the vector space is structured. As with numerous qualitative tasks, interpreting vectors relies on human expertise to evaluate and is highly dependent on the data at hand. It is rather straightforward to pick dimensions or words from the embedding space and analyze them with prior knowledge of semantics and syntax. Interpretability can also be leveraged with a more exploratory approach in mind when we ignore part of a network structure. For instance, we might wish to explore economical or biological networks with the goal of uncovering their structure.

\section{Conclusion}
\label{sec:conclusion}
Vector representations from networks or text have permitted tremendous progress in the exploitation and exploration of data. Thanks to increasingly powerful computational resources, we have seen the emergence of methods able to ingest ever larger amounts of data. The main limitations of these methods are the ecological impact of training large models on huge amounts of data, and also providing \textit{black-box} models that cannot be interpreted and audited. 

To circumvent these issues, we introduced a new framework deriving interpretable vector representations from networks and demonstrated the philosophy behind our framework on an airport network of the United States. The \emph{Lower Dimension Bipartite Graph Framework} (\LDBGF) we introduced aims to alleviate these shortcomings by projecting a network to a bipartite form and use the relationship between the vertices and the entities in the bipartite graph to compress information in vectors. Since \LDBGF is a framework, we propose in this paper two implementations: \SINrNR and \SINrMF. Both of these methods rely on community detection with the \louvain algorithm to project a network into a bipartite vertex-community relationship structure. Using this projection, \SINrNR uses the participation of each community to the degree of each vertex to derive a vector. In the case of \SINrMF, we aim to find the transition matrix between the network's adjacency matrix and the community-membership matrix obtained after community detection. Our method is applicable to any data that can be represented as an undirected network. 

For the purpose of this paper, we relied on classical networks representing citations, email exchanges, co-authorship and connections on social networks. We also carried out experiments on word co-occurrence networks extracted from large collections of curated documents. We evaluated our \SINrNR and \SINrMF approaches on classical quantitative evaluation tasks in each domain, namely \nlp and Network Science. We started by evaluating run time, \SINrNR is significantly faster than other graph embedding methods implemented in the same programming language. We then thoroughly evaluated the performances of \SINrNR and \SINrMF on three levels of organization of the networks: microscopic, mesoscopic and macroscopic levels. 

We started by probing microscopic-level information in networks with the classical \emph{link prediction} task. \SINrNR and \SINrMF are good representations to predict existing links and are close to other methods designed specifically for the task. We also demonstrated that accuracy on \emph{Link Prediction} can be optimized if we tune the number of dimensions for our models depending on the size of the network. Still at the microscopic-level, we also evaluated the capacity to predict degree of vertices from vectors. On this task, \SINrNR is undoubtedly the best performing model across the board. However, \SINrMF vectors are subpar in predicting vertex degree. \SINrNR success is probably because measuring the participation of a community to the degree of a vertex is a good precursor for predicting the degree of a vertex. Another vertex-level or micro-level characteristic is the clustering coefficient of a vertex. Once again, we fitted a regression model to the clustering coefficient of our networks' vertices. Unlike degree prediction, we found clustering coefficient are harder to predict. Overall, we found that none of the considered approaches properly predict clustering coefficient. However, \SINrNR still managed not to show massive discrepancies in results across all networks used, contrary to its competitors. 

Stepping back from the microscopic-level, we investigated the capacity of \SINrNR and \SINrMF to embed meso-scale information. For some networks in which we have ground truth community structures, we tried to reconstruct the partition in communities by clustering vertices from their embedding vector. What we have found that, although it is a complex task in an unsupervised setting, \SINrNR is on par with leading models on this task when \SINrMF cannot provide useful information. Switching from unsupervised to supervised community classification with the help of a classifier, we observed lower results for our approaches. Upon closer inspection, tweaking the number of communities and thus the dimension of vectors seems to increase classification performances. Overall, meso-scale information is relatively well embedded from the network into the vector representation.

Ascending to the highest level of information organization, we reach the macroscopic-level. Macro-scale information covers the whole network structure. To evaluate the extent of the macro-scale information in \SINrNR and \SINrMF, we selected the \emph{PageRank} score and tried to predict it for each vertex in the network. \emph{PageRank} is macroscopic in the sense that in a connected graph, every vertex has, to some extent, an influence on the score of each other vertex. \SINrNR manages to include useful information in its embedding to predict \emph{PageRank} scores almost perfectly. However, \SINrMF is outdone by \SINrNR and other baselines. 

Since text can also be represented using co-occurrence networks, we digress from our original subject of study to show the versatility of \SINrNR and its abilities to derive quality word embeddings. We employ two quantitative evaluations that are well defined in \nlp. \emph{Word similarity} evaluation is the first of our two quantitative experiments which measures the extent to which a model mimics similarities between words for humans. \SINrNR's results on \emph{word similarity} demonstrate that although originally a graph embedding model, it is a good fit for word embeddings. Furthermore, it has similar results to \FSPINE, an interpretable word embedding method. Our second experiment on text is analogous to community prediction, instead of communities, we try to cluster a subset of words into ontological categories. \emph{Concept categorization} results show that \SINrNR performs as well as \FSPINE and very similarly to \FWtoV.

Stable word representations are desirable in high stake applications, and guarantee that over the course of multiple runs, the output remains coherent. We thus aimed to answer two questions. How stable is our model over the course of multiple runs? Does it provide similar representations each time? We studied stability in two ways. First, we computed the variation in community structures across instances of \SINrNR. The results show that \louvain's community detection in the case of this approach uncovers roughly the same partition in communities each time. We also investigated the variation in neighborhood for models extracted from our word co-occurrence networks. An interesting finding is that \SINrNR presents very low variation for different runs on the same network, whereas \FSPINE's words neighborhoods change significantly.

We have seen through these quantitative evaluations that \SINrNR approaches can be good contenders on classical evaluation tasks. Although necessary, performance is not the only criterion we are interested in. Interpretability is a property that we encourage for sensitive applications of embedding methods, it opens up opportunities to understand models' internal structure. We demonstrated, through a human annotated \emph{word intrusion detection} evaluation that \SINrNR performs as well as \FSPINE, the state-of-the-art approach for word embedding interpretability. Despite these encouraging results, there is still a long way to go for perfect interpretability of models by humans. Through visualizations, we showed how dimensions can be interpreted relying just on the content and the values of vectors. These visualizations show the potential of \SINrNR to embed polysemy of words appearing in various contexts. Furthermore, we also showed that close neighbors share similar dimensions of the embedding space and that sense is modeled through a subset of dimensions from the vector space.

With all these experiments, we have thoroughly tested two implementations of \LDBGF: \SINrNR and \SINrMF. Although interpretable models can underperform on some tasks, yet, \SINrNR seems to be a well-rounded method to embed data. One could think that interpretability would come at the cost of performance but in most cases, \SINrNR and less often, \SINrMF are on par with baseline methods. 

These performances open the way for further developments within the framework. The first one would be to automatize interpretability evaluations, similarly to what was done by \textit{Lau et al.}~\cite{lau_machine_2014} for topic modeling. So far, our approaches do not account for directedness in networks. More specifically, for the text modality that we have studied, allowing \SINrNR to work with directed graphs would enable the possibility for syntactic word embeddings based on a syntactic dependency graph. Furthermore, we would like to handle temporal networks with \SINrNR and provide diachronic embeddings. The temporal word embeddings would benefit from the interpretability and stability characteristics of \SINrNR, while opening new opportunities to follow and analyze phenomenons through time. This is especially desirable for applications in semantic drift detection. Meanwhile, our methods have been made available to the public to learn and interpret representations, as well as collaborate towards more efficient and interpretable embedding methods.

\section*{Funding}

This work was supported by Agence Nationale de la Recherche (ANR) through the DIGING project [ANR-21-CE23-0010]. A CC-BY public copyright license has been applied by the authors to the present document and will be applied to all subsequent versions up to the Author Accepted Manuscript arising from this submission, in accordance with the grant’s open access conditions.
\begin{center}
    \includegraphics[width=.15\textwidth]{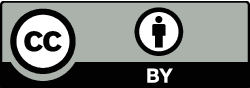}
\end{center}

\bibliographystyle{comnet}
\bibliography{bibliography}

\end{document}

%% file: figures/tikz/ldbgf/full_graph.tikz
\begin{tikzpicture}[scale=.9]
   \GraphInit[vstyle=Normal]
   \renewcommand*{\VertexLineWidth}{2pt}
   \SetGraphUnit{2}
   \Vertex[x=-1,y=0]{0}
   \Vertex[x=1.5, y=0.5]{1}
   \Vertex[x=0.5, y=-1.5]{2}
   \Vertex[x=3,  y=-1.5]{3}
   \Vertex[x=5, y =-1]{4}
   \Vertex[x=6, y=0.5]{5}
   \Vertex[x=8, y=0]{6}
   \Vertex[x=7, y=-1.5]{7}
  \Edges(0,1,3,2,0,3,2,1)
  \Edge(1)(2)
  \Edge(3)(4)
  \Edges(4,5,6,7,4, 5, 7)

\draw[rotate=-15, color=Burgundy, fill = Burgundy, fill opacity = 0.1] (1.2,-0.25) ellipse (2.8cm and 1.8cm);
\draw[rotate=14, color=cherryblossompink, fill = cherryblossompink, fill opacity = 0.2] (3.55,-2.2) ellipse (1.6cm and 0.8cm);
\draw[rotate=20, color=celadon, fill = celadon, fill opacity = 0.2] (5.3,-2.8) ellipse (1.5cm and 2.1cm);
\draw[rotate=23, color=teal, fill = teal, fill opacity = 0.2] (6.3,-3.2) ellipse (1.5cm and 2.1cm);
\node[] at (0, 0.8) {B0};
\node[] at (4.2, -1.7) {B1};
\node[] at (5, 0.3) {B2};
\node[] at (8, -1.5) {B3};
\end{tikzpicture}


%% file: figures/tikz/ldbgf/full_graph_adjacency.tex
\begin{tabular}{l|llllllll}
$\mathcal{A}$ & \textit{0} & \textit{1} & \textit{2} & \textit{3} & \textit{4} & \textit{5} & \textit{6} & \textit{7} \\ \hline
\textit{0} & 0 & 1 & 1 & 1 & 0 & 0 & 0 & 0 \\
\textit{1} & 1 & 0 & 1 & 1 & 0 & 0 & 0 & 0 \\
\textit{2} & 1 & 1 & 0 & 1 & 0 & 0 & 0 & 0 \\
\textit{3} & 1 & 1 & 1 & 0 & 1 & 0 & 0 & 0 \\
\textit{4} & 0 & 0 & 0 & 1 & 0 & 1 & 0 & 1 \\
\textit{5} & 0 & 0 & 0 & 0 & 1 & 0 & 1 & 1 \\
\textit{6} & 0 & 0 & 0 & 0 & 0 & 1 & 0 & 1 \\
\textit{7} & 0 & 0 & 0 & 0 & 1 & 1 & 1 & 0
\end{tabular}


%% file: figures/tikz/ldbgf/bipartite_cliques.tikz
        \begin{tikzpicture}[scale=.6]
  
   \SetGraphUnit{3}
   \renewcommand*{\VertexLineWidth}{2pt}
   \SetVertexNormal[FillColor = Burgundy!20]
    \Vertex[x=3,y=2.5]{B0}
    \SetVertexNormal[FillColor = cherryblossompink!20]
    \Vertex[x=6, y=2.5]{B1}
    \SetVertexNormal[FillColor = celadon!20]
    \Vertex[x=9, y=2.5]{B2}
    \SetVertexNormal[FillColor = teal!20]
    \Vertex[x=12, y=2.5]{B3}
    
    \SetGraphUnit{2}
    \SetVertexNormal[]
    \Vertices[x=0, y=-2]{line}{0,1,2,3}
    \Vertices[x=8, y=-2]{line}{4,5,7,6}
    \Edges(B0,0,B0,1,B0,2, B0, 3)
    \Edges(B1,3,B1,4)
    \Edges(B2,4,B2,5, B2, 7)
    \Edges(B3,5,B3,6,B3, 7)

\end{tikzpicture}


%% file: figures/tikz/ldbgf/bipartite_adjacency.tex
{\small
\begin{tabular}{l|uynt}
$\mathcal{A}^{\prime}$ & \textit{B0} & \textit{B1} & \textit{B2} & \textit{B3}  \\ \hline
\textit{0} & 1  & 0  & 0  & 0    \\
\textit{1} & 1  & 0  & 0  & 0    \\
\textit{2} & 1  & 0  & 0  & 0    \\
\textit{3} & 1  & 1  & 0  & 0    \\
\textit{4} & 0  & 1  & 1  & 0    \\
\textit{5} & 0  & 0  & 1  & 1    \\
\textit{6} & 0  & 0  & 0  & 1    \\
\textit{7} & 0  & 0  & 1  & 1  
\end{tabular}%
}


%% file: figures/tikz/sinr/full_graph_communities.tikz
\begin{tikzpicture}[scale=.9]
   \SetGraphUnit{2}
   \renewcommand*{\VertexLineWidth}{2pt}

   \SetVertexNormal[FillColor = cherryblossompink!20]
   \Vertex[x=-1,y=0]{0}
   \Vertex[x=1.5, y=0.5]{1}
   \Vertex[x=0.5, y=-1.5]{2}
   \Vertex[x=3,  y=-1.5]{3}
   \SetVertexNormal[FillColor = celadon!20]
   \Vertex[x=5, y =-1]{4}
   \Vertex[x=6, y=0.5]{5}
   \Vertex[x=8, y=0]{6}
   \Vertex[x=7, y=-1.5]{7}
  \Edges(0,1,3,2,0,2,1,0,3)
  \Edge(1)(2)
  \Edge(3)(4)
  \Edges(4,5,6,7,4,5,7)
\end{tikzpicture}

%% file: figures/tikz/sinr/bipartite_communities.tikz
\tikzset{LabelStyle/.style  = {}}

\begin{tikzpicture}[scale=1]
   \SetGraphUnit{2}
   \renewcommand*{\VertexLineWidth}{2pt}
   \SetVertexNormal[FillColor = cherryblossompink!20]
    \Vertices[x=0, y=-2]{line}{0,1,2,3}
    \Vertex[x=3.5,y=0]{C0}
    \SetVertexNormal[FillColor = celadon!20]
    \Vertices[x=8, y=-2]{line}{4,5,6,7}
    \Vertex[x=10, y=0]{C1}
    \Edge[label=$1$](C0)(0)
    \Edge[label=$1$](C0)(1)
    \Edge[label=$1$](C0)(2)
    \Edge[label=$0.75$](C0)(3)
    \Edge[label=$0.33$](C0)(4)
    \Edge[label=$0.25$](C1)(3)
    \Edge[label=$0.66$](C1)(4)
    \Edge[label=$1$](C1)(5)
    \Edge[label=$1$](C1)(6)
    \Edge[label=$1$](C1)(7)
\end{tikzpicture}

%% file: figures/tikz/sinr/sinr_emb_matrix.tex
\begin{tabular}{c|yn}
           & \textit{C0} & \textit{C1} \\ \hline
\textit{0} &  1           &  0          \\
\textit{1} &  1           &  0          \\
\textit{2} &  1           &  0          \\
\textit{3} &  0.75        &  0.25        \\
\textit{4} &  0.33        &  0.66        \\
\textit{5} &  0           &  1          \\
\textit{6} &  0           &  1          \\
\textit{7} &  0           &  1         
\end{tabular}%

%% file: Author_tex.bbl
\begin{thebibliography}{00}

\bibitem{almuhareb_concept_2005}
Almuhareb, A. {\&} Poesio, M. (2005)  Concept learning and categorization from the web. In {\em proceedings of the annual meeting of the Cognitive Science society}, volume~27.

\bibitem{baroni_dont_2014}
Baroni, M., Dinu, G. {\&} Kruszewski, G. (2014)  Don't count, predict! {A} systematic comparison of context-counting vs. context-predicting semantic vectors. In {\em ACL}, pages 238--247.

\bibitem{baroni_how_2011}
Baroni, M. {\&} Lenci, A. (2011)  How we {BLESSed} distributional semantic evaluation. {\em GEometrical Models of Natural Language Semantics}, pages 1--10.

\bibitem{belkin_laplacian_2001}
Belkin, M. {\&} Niyogi, P. (2001)  Laplacian eigenmaps and spectral techniques for embedding and clustering. {\em Advances in neural information processing systems}, \textbf{14}.

\bibitem{bhowmick_louvainne_2020}
Bhowmick, A.~K., Meneni, K., Danisch, M., Guillaume, J.-L. {\&} Mitra, B. (2020)  {LouvainNE}: {Hierarchical} {Louvain} {Method} for {High} {Quality} and {Scalable} {Network} {Embedding}. In {\em WSDM}, pages 43--51.

\bibitem{blondel_fast_2008}
Blondel, V.~D., Guillaume, J.~L., Lambiotte, R. {\&} Lefebvre, E. (2008)  Fast unfolding of communities in large networks. {\em Journal of Statistical Mechanics: Theory and Experiment}.
arXiv: 0803.0476.

\bibitem{bohlin_community_2014}
Bohlin, L., Edler, D., Lancichinetti, A. {\&} Rosvall, M. (2014)  Community detection and visualization of networks with the map equation framework. {\em Measuring scholarly impact: methods and practice}, pages 3--34.

\bibitem{bojanowski_enriching_2017}
Bojanowski, P., Grave, E., Joulin, A. {\&} Mikolov, T. (2017)  Enriching {Word} {Vectors} with {Subword} {Information}. {\em Transactions of the Association for Computational Linguistics}.

\bibitem{brin_anatomy_1998}
Brin, S. {\&} Page, L. (1998)  The anatomy of a large-scale hypertextual {Web} search engine. {\em Computer Networks and ISDN Systems}, \textbf{30}(1-7), 107--117.

\bibitem{brochier_global_2019}
Brochier, R., Guille, A. {\&} Velcin, J. (2019)  Global {Vectors} for {Node} {Representations}. In {\em The {World} {Wide} {Web} {Conference}}, pages 2587--2593.
arXiv:1902.11004 [cs].

\bibitem{broniatowski_psychological_2021}
Broniatowski, D.~A.  et~al. (2021)  Psychological foundations of explainability and interpretability in artificial intelligence. {\em NIST, Tech. Rep}.

\bibitem{brown_language_2020}
Brown, T., Mann, B., Ryder, N., Subbiah, M., Kaplan, J.~D., Dhariwal, P., Neelakantan, A., Shyam, P., Sastry, G., Askell, A.  et~al. (2020)  Language models are few-shot learners. {\em Neurips}, \textbf{33}, 1877--1901.

\bibitem{bruni_multimodal_2014}
Bruni, E., Tran, N.~K. {\&} Baroni, M. (2014)  Multimodal {Distributional} {Semantics}. {\em Journal of Artificial Intelligence Research}, \textbf{49}, 1--47.

\bibitem{cao_grarep_2015}
Cao, S., Lu, W. {\&} Xu, Q. (2015)  {GraRep}: {Learning} {Graph} {Representations} with {Global} {Structural} {Information}. In {\em CIKM}, pages 891--900.

\bibitem{cao_deep_2016}
Cao, S., Lu, W. {\&} Xu, Q. (2016)  Deep neural networks for learning graph representations. In {\em Proceedings of the AAAI conference on artificial intelligence}, volume~30.

\bibitem{chakraborty_metadata_2018}
Chakraborty, T., Cui, Z. {\&} Park, N. (2018)  Metadata vs. {Ground}-truth: {A} {Myth} behind the {Evolution} of {Community} {Detection} {Methods}. In {\em {WWW}'18}, pages 45--46, Lyon, France. ACM Press.

\bibitem{chang_reading_2009}
Chang, J., Gerrish, S., Wang, C., Boyd-graber, J. {\&} Blei, D. (2009)  Reading {Tea} {Leaves}: {How} {Humans} {Interpret} {Topic} {Models}. In {\em Neurips}, volume~22.

\bibitem{chen_unsupervised_2008}
Chen, J., Zaïane, O.~R. {\&} Goebel, R. (2008)  An Unsupervised Approach to Cluster Web Search Results Based on Word Sense Communities. In {\em International Conference on Web Intelligence and Intelligent Agent Technology}, volume~1, pages 725--729.

\bibitem{choudhary2022survey}
Choudhary, M., Laclau, C. {\&} Largeron, C. (2022)  A survey on fairness for machine learning on graphs. {\em arXiv preprint arXiv:2205.05396}.

\bibitem{bnc_oxford}
Consortium, B. (2007)  British National Corpus, {XML} edition. Oxford Text Archive.

\bibitem{dao_community_2017}
Dao, V.-L., Bothorel, C. {\&} Lenca, P. (2017)  Community detection methods can discover better structural clusters than ground-truth communities. In {\em {ASoNAM}'17}, pages 395--400, Sydney Australia. ACM.

\bibitem{devlin_bert_2018}
Devlin, J., Chang, M.-W., Lee, K. {\&} Toutanova, K. (2019)  {BERT}: Pre-training of Deep Bidirectional Transformers for Language Understanding. In {\em North {A}merican Chapter of the Association for Computational Linguistics}, pages 4171--4186.

\bibitem{dugue_bringing_2019}
Dugu{\'e}, N., Lamirel, J.-C. {\&} Perez, A. (2019)  Bringing a feature selection metric from machine learning to complex networks. In {\em Complex Networks and Their Applications VII}, pages 107--118.

\bibitem{duong_interpretable_2019}
Duong, C.~T., Nguyen, Q. V.~H. {\&} Aberer, K. (2019)  Interpretable node embeddings with mincut loss. In {\em Learning and Reasoning with Graph-Structured Representations Workshop-ICML}.

\bibitem{erdoos1988clique}
Erd{\"o}os, P., Faudree, R. {\&} Ordman, E.~T. (1988)  Clique partitions and clique coverings. {\em Discrete Mathematics}, \textbf{72}(1-3), 93--101.

\bibitem{esslli_dataset_2008}
ESSLLI (2008)  Shared Tasks from the {ESSLLI} 2008 Workshop. Data \& Description : http://wordspace.collocations.de/doku.php/data:esslli2008:start.

\bibitem{faruqui_sparse_2015}
Faruqui, M., Tsvetkov, Y., Yogatama, D., Dyer, C. {\&} Smith, N.~A. (2015)  Sparse {Overcomplete} {Word} {Vector} {Representations}. In {\em ACL}, pages 1491--1500.

\bibitem{finkelstein_placing_2001}
Finkelstein, L., Gabrilovich, E., Matias, Y., Rivlin, E., Solan, Z., Wolfman, G. {\&} Ruppin, E. (2001)  Placing search in context: {The} concept revisited. {\em WWW}, pages 406--414.

\bibitem{fortunato_resolution_2007}
Fortunato, S. {\&} Barthélemy, M. (2007)  Resolution limit in community detection. {\em Proceedings of the National Academy of Sciences}, \textbf{104}(1), 36--41.

\bibitem{garg2018word}
Garg, N., Schiebinger, L., Jurafsky, D. {\&} Zou, J. (2018)  Word embeddings quantify 100 years of gender and ethnic stereotypes. {\em Proceedings of the National Academy of Sciences}, \textbf{115}(16), E3635--E3644.

\bibitem{gefen2017vector}
Gefen, A., Algee-Hewitt, M.~A., McClure, D., Glorieux, F., Reboul, M., Porter, J. {\&} Riguet, M. (2017)  Vector based measure of semantic shifts across different cultural corpora as a proxy to comparative history of ideas. {\em JADH 2017}, page~12.

\bibitem{GN02}
Girvan, M. {\&} Newman, M.~E. (2002)  Community structure in social and biological networks. {\em Proceedings of the national academy of sciences}, \textbf{99}(12), 7821--7826.

\bibitem{grover_node2vec_2016}
Grover, A. {\&} Leskovec, J. (2016)  node2vec: {Scalable} {Feature} {Learning} for {Networks}. arXiv:1607.00653 [cs, stat].

\bibitem{GL04}
Guillaume, J.-L. {\&} Latapy, M. (2004)  Bipartite structure of all complex networks. {\em Information processing letters}, \textbf{90}(5), 215--221.

\bibitem{hamilton2016cultural}
Hamilton, W.~L., Leskovec, J. {\&} Jurafsky, D. (2016)  Cultural shift or linguistic drift? comparing two computational measures of semantic change. In {\em EMNLP}, page 2116.

\bibitem{harris_distributional_1954}
Harris, Z.~S. (1954)  Distributional structure. {\em Word}, \textbf{10}(2-3), 146--162.
Publisher: Taylor \& Francis.

\bibitem{huang_improving_2012}
Huang, E.~H., Socher, R., Manning, C.~D. {\&} Ng, A.~Y. (2012)  Improving word representations via global context and multipleword prototypes. In {\em ACL}.

\bibitem{kim2023race}
Kim, M., Kim, J. {\&} Johnson, K. (2023)  Race, Gender, and Age Biases in Biomedical Masked Language Models. In {\em ACL}, pages 11806--11815.

\bibitem{lambiotte2013multi}
Lambiotte, R. (2013)  Multi-scale modularity and dynamics in complex networks. In {\em Dynamics On and Of Complex Networks, Volume 2: Applications to Time-Varying Dynamical Systems}, pages 125--141.

\bibitem{lancichinetti_detecting_2009}
Lancichinetti, A., Fortunato, S. {\&} Kertész, J. (2009)  Detecting the overlapping and hierarchical community structure in complex networks. {\em New Journal of Physics}, \textbf{11}(3), 033015.

\bibitem{lancichinetti_characterize_2010}
Lancichinetti, A., Kivelä, M., Saramäki, J. {\&} Fortunato, S. (2010)  Characterizing the Community Structure of Complex Networks. {\em PLOS ONE}, \textbf{5}(8), 1--8.

\bibitem{lancichinetti_OSLOM_2011}
Lancichinetti, A., Radicchi, F., Ramasco, J.~J. {\&} Fortunato, S. (2011)  Finding Statistically Significant Communities in Networks. {\em PLOS ONE}, \textbf{6}(4), 1--18.

\bibitem{lannelongue_green_2021}
Lannelongue, L., Grealey, J. {\&} Inouye, M. (2021)  Green {Algorithms}: {Quantifying} the {Carbon} {Footprint} of {Computation}. {\em Advanced Science}, \textbf{8}(12), 2100707.

\bibitem{lau_machine_2014}
Lau, J.~H., Newman, D. {\&} Baldwin, T. (2014)  Machine {Reading} {Tea} {Leaves}: {Automatically} {Evaluating} {Topic} {Coherence} and {Topic} {Model} {Quality}. In {\em EACL}, pages 530--539.

\bibitem{lee2014community}
Lee, C. {\&} Cunningham, P. (2014)  Community detection: effective evaluation on large social networks. {\em Journal of Complex Networks}, \textbf{2}(1), 19--37.

\bibitem{lee99}
Lee, D.~D. {\&} Seung, H.~S. (1999)  Learning the parts of objects by nonnegative matrix factorization. {\em Nature}, \textbf{401}, 788--791.

\bibitem{levy_neural_2014}
Levy, O. {\&} Goldberg, Y. (2014)  Neural Word Embedding as Implicit Matrix Factorization. In {\em Neurips}, volume~27.

\bibitem{LGD15}
Levy, O., Goldberg, Y. {\&} Dagan, I. (2015)  Improving distributional similarity with lessons learned from word embeddings. {\em ACL}, \textbf{3}, 211--225.

\bibitem{liu_online_2016}
Liu, F., Yang, X., Guan, N. {\&} Yi, X. (2016)  Online graph regularized non-negative matrix factorization for large-scale datasets. {\em Neurocomputing}, \textbf{204}, 162--171.

\bibitem{liu_roberta_2019}
Liu, Y., Ott, M., Goyal, N., Du, J., Joshi, M., Chen, D., Levy, O., Lewis, M., Zettlemoyer, L. {\&} Stoyanov, V. (2019)  {RoBERTa}: {A} {Robustly} {Optimized} {BERT} {Pretraining} {Approach}. arXiv:1907.11692 [cs].

\bibitem{lundberg_unified_2017}
Lundberg, S.~M. {\&} Lee, S.-I. (2017)  A unified approach to interpreting model predictions. {\em Advances in neural information processing systems}, \textbf{30}.

\bibitem{makarov_survey_2021}
Makarov, I., Kiselev, D., Nikitinsky, N. {\&} Subelj, L. (2021)  Survey on graph embeddings and their applications to machine learning problems on graphs. {\em PeerJ Computer Science}, \textbf{7}, e357.

\bibitem{mikolov_efficient_2013}
Mikolov, T., Chen, K., Corrado, G. {\&} Dean, J. (2013)  Efficient estimation of word representations in vector space. In {\em ICLR}.
arXiv: 1301.3781.

\bibitem{miller_wordnet_1995}
Miller, G.~A. (1995)  WordNet: A Lexical Database for English. {\em Commun. ACM}, \textbf{38}(11), 39–41.

\bibitem{murphy_learning_2012}
Murphy, B., Talukdar, P.~P. {\&} Mitchell, T. (2012)  Learning effective and interpretable semantic models using non-negative sparse embedding. In {\em COLING}, pages 1933--1950.

\bibitem{oanc_corpus}
Nancy~Ide, Randi~Reppen, K.~S. (2011)  The Open ANC (OANC). {ORTOLANG}.

\bibitem{neelakantan_text_nodate}
Neelakantan, A., Xu, T., Puri, R., Radford, A., Han, J.~M., Tworek, J., Yuan, Q., Tezak, N., Kim, J.~W., Hallacy, C.  et~al. (2022)  Text and code embeddings by contrastive pre-training. {\em arXiv preprint arXiv:2201.10005}.

\bibitem{neveol2022french}
N{\'e}v{\'e}ol, A., Dupont, Y., Bezan{\c{c}}on, J. {\&} Fort, K. (2022)  French CrowS-Pairs: Extending a challenge dataset for measuring social bias in masked language models to a language other than English. In {\em ACL}, pages 8521--8531.

\bibitem{osgood_measurement_1957}
Osgood, C.~E., Suci, G.~J. {\&} Tannenbaum, P.~H. (1957) {\em The measurement of meaning.}
The measurement of meaning. Univer. Illinois Press, Oxford, England.
Pages: 342.

\bibitem{ou_asymmetric_2016}
Ou, M., Cui, P., Pei, J., Zhang, Z. {\&} Zhu, W. (2016)  Asymmetric {Transitivity} {Preserving} {Graph} {Embedding}. In {\em SIGKDD}, pages 1105--1114.

\bibitem{PALMER1977441}
Palmer, S.~E. (1977)  Hierarchical structure in perceptual representation. {\em Cognitive Psychology}, \textbf{9}(4), 441--474.

\bibitem{panigrahi_word2sense_2019}
Panigrahi, A., Simhadri, H.~V. {\&} Bhattacharyya, C. (2019)  {Word2Sense}: {Sparse} {Interpretable} {Word} {Embeddings}. In {\em ACL}, pages 5692--5705.

\bibitem{patterson_carbon_nodate}
Patterson, D., Gonzalez, J., Le, Q., Liang, C., Munguia, L.-M., Rothchild, D., So, D., Texier, M. {\&} Dean, J. (2021)  Carbon emissions and large neural network training. {\em arXiv preprint arXiv:2104.10350}.

\bibitem{peel2017ground}
Peel, L., Larremore, D.~B. {\&} Clauset, A. (2017)  The ground truth about metadata and community detection in networks. {\em Science advances}, \textbf{3}(5), e1602548.

\bibitem{pennington_glove_2014}
Pennington, J., Socher, R. {\&} Manning, C.~D. (2014)  Glove: Global vectors for word representation. In {\em EMNLP}, pages 1532--1543.

\bibitem{perozzi_deepwalk_2014}
Perozzi, B., Al-Rfou, R. {\&} Skiena, S. (2014)  {DeepWalk}: {Online} {Learning} of {Social} {Representations}. {\em SIGKDD}, pages 701--710.
arXiv: 1403.6652.

\bibitem{perozzi_dont_2017}
Perozzi, B., Kulkarni, V., Chen, H. {\&} Skiena, S. (2017)  Don't {Walk}, {Skip}! {Online} {Learning} of {Multi}-scale {Network} {Embeddings}. In {\em {ASONAM}}, pages 258--265.

\bibitem{pierrejean:tel-02628954}
Pierrejean, B. (2020) {\em {Qualitative Evaluation of Word Embeddings: Investigating the Instability in Neural-Based Models}}.
PhD thesis, {Universit{\'e} Toulouse 2 - Jean Jaur{\`e}s}.

\bibitem{prouteau_sinr_2021}
Prouteau, T., Connes, V., Dugué, N., Perez, A., Lamirel, J.-C., Camelin, N. {\&} Meignier, S. (2021)  {SINr}: {Fast} {Computing} of {Sparse} {Interpretable} {Node} {Representations} is not a {Sin}!. In {\em IDA}, pages 325--337.

\bibitem{prouteau:hal-03770444}
Prouteau, T., Dugu{\'e}, N., Camelin, N. {\&} Meignier, S. (2022)  {Are Embedding Spaces Interpretable? Results of an Intrusion Detection Evaluation on a Large French Corpus}. In {\em {LREC 2022}}, page 4414–4419.

\bibitem{raffel_exploring_2020}
Raffel, C., Shazeer, N., Roberts, A., Lee, K., Narang, S., Matena, M., Zhou, Y., Li, W. {\&} Liu, P.~J. (2020)  Exploring the {Limits} of {Transfer} {Learning} with a {Unified} {Text}-to-{Text} {Transformer}. arXiv:1910.10683.

\bibitem{raghavan_near_2007}
Raghavan, U.~N., Albert, R. {\&} Kumara, S. (2007)  Near linear time algorithm to detect community structures in large-scale networks. {\em Physical Review E - Statistical, Nonlinear, and Soft Matter Physics}.
arXiv: 0709.2938.

\bibitem{ribeiro_why_2016}
Ribeiro, M.~T., Singh, S. {\&} Guestrin, C. (2016)  "{Why} {Should} {I} {Trust} {You}?": {Explaining} the {Predictions} of {Any} {Classifier}. arXiv:1602.04938 [cs, stat].

\bibitem{roweis_nonlinear_2000}
Roweis, S.~T. {\&} Saul, L.~K. (2000)  Nonlinear dimensionality reduction by locally linear embedding. {\em science}, \textbf{290}(5500), 2323--2326.
Publisher: American Association for the Advancement of Science.

\bibitem{rozemberczki_gemsec_2019}
Rozemberczki, B., Davies, R., Sarkar, R. {\&} Sutton, C. (2020)  GEMSEC: graph embedding with self clustering. .

\bibitem{rubenstein_contextual_1965}
Rubenstein, H. {\&} Goodenough, J.~B. (1965)  Contextual correlates of synonymy. {\em Communications of the ACM}, \textbf{8}(10), 627--633.

\bibitem{rudin_stop_2019}
Rudin, C. (2019)  Stop explaining black box machine learning models for high stakes decisions and use interpretable models instead. {\em Nature machine intelligence}, \textbf{1}(5), 206--215.

\bibitem{serra_interpreting_2021}
Serra, G., Xu, Z., Niepert, M., Lawrence, C., Tino, P. {\&} Yao, X. (2021)  Interpreting {Node} {Embedding} with {Text}-labeled {Graphs}. In {\em IJCNN}, pages 1--8.

\bibitem{shi_normalized_2000}
Shi, J. {\&} Malik, J. (2000)  Normalized {Cuts} and {Image} {Segmentation}. {\em IEEE Transactions on pattern analysis and machine intelligence}, \textbf{22}(8).

\bibitem{sinha2019systematic}
Sinha, A., Cazabet, R. {\&} Vaudaine, R. (2019)  Systematic biases in link prediction: comparing heuristic and graph embedding based methods. In {\em Complex Networks}, pages 81--93.

\bibitem{StaudtSM14}
Staudt, C., Sazonovs, A. {\&} Meyerhenke, H. (2014)  NetworKit: An Interactive Tool Suite for High-Performance Network Analysis. {\em CoRR}, \textbf{abs/1403.3005}.

\bibitem{strubell_energy_2020}
Strubell, E., Ganesh, A. {\&} McCallum, A. (2020)  Energy and policy considerations for modern deep learning research. In {\em Proceedings of the AAAI conference on artificial intelligence}, volume~34, pages 13693--13696.

\bibitem{subramanian_spine_2018}
Subramanian, A., Pruthi, D., Jhamtani, H., Berg-Kirkpatrick, T. {\&} Hovy, E. (2018)  {SPINE}: {SParse} {Interpretable} {Neural} {Embeddings}. In {\em Thirty-{Second} {AAAI} {Conference} on {Artificial} {Intelligence}}.

\bibitem{tang_line_2015}
Tang, J., Qu, M., Wang, M., Zhang, M., Yan, J. {\&} Mei, Q. (2015)  {LINE}: {Large}-scale {Information} {Network} {Embedding}. In {\em WWW}, pages 1067--1077.

\bibitem{tenenbaum_global_2000}
Tenenbaum, J.~B., Silva, V.~d. {\&} Langford, J.~C. (2000)  A {Global} {Geometric} {Framework} for {Nonlinear} {Dimensionality} {Reduction}. {\em Science}, \textbf{290}(5500), 2319--2323.

\bibitem{tian2014probabilistic}
Tian, F., Dai, H., Bian, J., Gao, B., Zhang, R., Chen, E. {\&} Liu, T.-Y. (2014)  A probabilistic model for learning multi-prototype word embeddings. In {\em COLING}, pages 151--160.

\bibitem{tsitsulin_verse_2018}
Tsitsulin, A., Mottin, D., Karras, P. {\&} Müller, E. (2018)  {VERSE}: {Versatile} {Graph} {Embeddings} from {Similarity} {Measures}. In {\em WWW}, pages 539--548.
arXiv:1803.04742 [cs].

\bibitem{wang_structural_2016}
Wang, D., Cui, P. {\&} Zhu, W. (2016)  Structural {Deep} {Network} {Embedding}. In {\em SIGKDD}, pages 1225--1234.

\bibitem{wang_community_2017}
Wang, X., Cui, P., Wang, J., Pei, J., Zhu, W. {\&} Yang, S. (2017)  Community {Preserving} {Network} {Embedding}. {\em Proceedings of the AAAI Conference on Artificial Intelligence}, \textbf{31}(1).

\bibitem{yang2015network}
Yang, C., Liu, Z., Zhao, D., Sun, M. {\&} Chang, E.~Y. (2015)  Network representation learning with rich text information.. In {\em IJCAI}, volume 2015, pages 2111--2117.

\end{thebibliography}
